\documentclass[11pt]{article}

\usepackage[a4paper, left=3.2cm,top=3.2cm]{geometry}
\usepackage{authblk}
\usepackage[onehalfspacing]{setspace}

\usepackage{tikz}
\usetikzlibrary{decorations.pathreplacing}

\usepackage{pgfplots}
\pgfplotsset{compat=1.10}
\usetikzlibrary{intersections}

\usepackage{siunitx}
\usepackage{booktabs}

\usepackage{amsmath,amsfonts}
\usepackage{algorithmic}
\usepackage{algorithm}
\usepackage{array}
\usepackage[caption=false,font=normalsize,labelfont=sf,textfont=sf]{subfig}
\usepackage{textcomp}
\usepackage{stfloats}
\usepackage{url}
\usepackage{verbatim}
\usepackage{graphicx}
\usepackage{cite}
\usepackage{cite}

\usepackage{mathtools}
\usepackage{enumitem}
\usepackage{bm}
\usepackage{amssymb}

\DeclareMathOperator*{\argmin}{arg\, min}
\DeclareMathOperator*{\argmax}{arg\, max}

\newcommand{\id}{\operatorname{Id}}
\newcommand{\ran}{\operatorname{ran}}

\newcommand{\N}{\mathbb{N}}
\newcommand{\R}{\mathbb{R}}

\newcommand{\C}{\mathbb{C}}

\newcommand\norm[1]{\lVert#1\rVert}

\newcommand\abs[1]{\lvert#1\rvert}

\newcommand{\paren}[1]{\left(#1\right)}               

\newcommand{\solspace}{\mathbb{L}}

\usepackage{amsthm}

\numberwithin{equation}{section}
\numberwithin{theorem}{section}

\newcommand{\Ao}{\mathcal{A}}
\newcommand{\Ro}{\mathcal{R}}
\newcommand{\Bo}{\mathcal{B}}

\newcommand{\Mo}{\mathcal{M}}
\newcommand{\No}{\mathcal{N}}
\newcommand{\Uo}{\mathcal{U}}
\newcommand{\Po}{\mathcal{P}}
\newcommand{\Fo}{\mathcal{F}}
\newcommand{\So}{\mathcal{S}}

\newcommand{\signal}{x}
\newcommand{\data}{y}

\newcommand{\risk}{\mathcal{L}}
\newcommand{\riskunc}{\mathcal{L}^\text{unc}}

\DeclareMathOperator{\ssim}{SSIM}
\DeclareMathOperator{\psnr}{PSNR}

\allowdisplaybreaks

\title{Uncertainty-Aware Null Space Networks for Data-Consistent Image Reconstruction}

\date{April 14, 2023}

\author{Christoph~Angermann}

\affil{VASCage -- Centre on Clinical Stroke Research\authorcr
6020 Innsbruck, Austria
 \authorcr E-mail:  \texttt{christoph.angermann@vascage.at}
 }

\author{Simon Göppel}

\affil{Department of Mathematics, University of Innsbruck\authorcr
Technikerstrasse 13, 6020 Innsbruck, Austria
 \authorcr E-mail:  \texttt{simon.goeppel@uibk.ac.at}
 }

\author{Markus Haltmeier}

\affil{Department of Mathematics, University of Innsbruck\authorcr
Technikerstrasse 13, 6020 Innsbruck, Austria
 \authorcr E-mail:  \texttt{markus.haltmeier@uibk.ac.at}
 }

\begin{document}

\maketitle

\begin{abstract}

Reconstructing an image from noisy and incomplete measurements is a central task in several image processing applications. In recent years, state-of-the-art reconstruction methods have been developed based on recent advances in deep learning. Especially for highly underdetermined problems, maintaining data consistency is a key goal. This can be achieved either by iterative network architectures or by a subsequent projection of the network reconstruction. However, for such approaches to be used in safety-critical domains such as medical imaging, the network reconstruction should not only provide the user with a reconstructed image, but also with some level of confidence in the reconstruction. In order to meet these two key requirements, this paper combines deep null-space networks with uncertainty quantification. Evaluation of the proposed method includes image reconstruction from undersampled Radon measurements on a toy CT dataset and accelerated MRI reconstruction on the fastMRI dataset. This work is the first approach to solving inverse problems that additionally models data-dependent uncertainty by estimating an input-dependent scale map, providing a robust assessment of reconstruction quality.


\end{abstract}

\section{Introduction}
\label{sec:intro}

Inverse problems arise in a variety of sciences and play a central role in many engineering applications \cite{engl1996regularization}. After discretization \cite{kaipio2006statistical,hansen2010discrete}, an inverse problem can be written in the form
	\begin{equation} \label{eq:ip}
		\data = \Ao \signal + \epsilon,
	\end{equation}
	where $\data, \epsilon \in \R^m$, $\signal\in \R^n$ and $\Ao \in \R^{m \times n}$. The goal is to recover the unknown signal $\signal$ from the knowledge of the noisy data $\data$ and the measurement operator $\Ao$. In the deterministic case, the  data perturbation $\epsilon$ is assumed to satisfy  $\norm{\epsilon}_2 \leq \delta$ for some noise level $\delta>0$. Inverse problems are characterized by non-uniqueness and sensitivity with respect to data perturbations. In particular, for applications with incomplete data, the reduced number of measurements results in a high dimensional set of potential solutions. In order to obtain reliable reconstructions, it is therefore necessary to apply suitable regularization techniques to the problem at hand, which address non-uniqueness and instability~\cite{engl1996regularization,scherzer2009variational}.
	
Prime examples of ill-posed inverse problems are found in image reconstruction, such as computed tomography (CT) and magnetic resonance imaging (MRI). Both of these techniques often encounter problems with limited data because measurements are only available for a limited subset. This is the case, for example, with limited-angle computed tomography, where angular projections are available only within a strict subset of the full angular range. These limitations may be due to the physical measurement setup, for example, and naturally occur in practical applications such as digital breast tomosynthesis, dental tomography or nondestructive testing. Due to the lack of available data, important features in the reconstructions can be obscured by artifacts caused by hard cut-offs in the measurement  domain. While the characterization of these artifacts has been thoroughly investigated \cite{frikel2013,frikel2016,borg2018}, the reliable correction of missing data is still a challenging task.  Unlike computed tomography, MRI takes measurements without exposing the patient to ionizing radiation. MRI has a variety of applications for disease detection, general treatment, and prognosis. To save cost and time, it is common practice in MRI to reduce the acquisition time by taking an incomplete number of measurements. Due to the resulting undersampling, the  patient image is no longer uniquely determined by the available data.  To overcome this limitation, methods of compressed sensing (CS) and $\ell^1$-regularization \cite{Candes2006,Donoho2006,grasmair2008sparse,Lustig2008,block2007undersampled,popilka2007} have been developed which allow accurate reconstruction from sparsely sampled data. As a result, CS and sparsity regularization became key tools in modern MRI imaging, and is still an active area of research.
	
In recent years, machine learning (ML) and neural networks (NN) have emerged as new paradigms for solving inverse problems  \cite{arridge2019,jin2017,li2020,ongie2020,wang2020deep,adler2017solving,altekruger2022}, in which a trainable NN is adjusted to the available dataset. The fitting process, also referred to as network training, can be treated both as a supervised or unsupervised learning task. Supervised training attempts to infer a cross-domain correspondence between measurements and ground truth data, while unsupervised training attempts to estimate an overall image distribution from which reconstructions can be sampled \cite{goodfellow2014,creswell2018}. Note that any inverse problem can be considered a supervised ML task, as long as a sufficient amount of data is available and the forward operator is known explicitly. In many cases, data-driven solvers significantly reduce reconstruction time while improving accuracy. However, given the highly nonlinear structure of a NN, it remains unclear how accurately a network processes the provided data. This is of particular importance for medical applications, where reliable reconstruction is essential. In particular, for problems with limited data, it is desirable to explicitly enforce data consistency in order to obtain reliable predictions. The difficult-to-interpret nature of NNs combined with the lack of theoretical guarantees still limits the practical applicability of DL-based solvers in clinical trials \cite{elad2017,genzel2022}. Null space networks that promote data-consistent reconstruction through architectural design have been introduced and analyzed in \cite{schwab2019,schwab2020big} and shown to result in a convergent regularization technique. 

Due to the difficult-to-interpret structure, it is often argued that NNs suffer from the so-called black-box paradigm \cite{buhrmester2021}.  In general, no interpretable connection can be made between the learned network parameters and the actual inter-domain relationship. As a result, network behavior is almost unpredictable in the presence of out-of-distribution (OOD) data, which are not included in the training set \cite{kendall2017,abdar2021}. OOD data occurs, for example, when the scanner's measurement quality suddenly degrades or unusual objects (such as tumors or metal parts in medical imaging) become visible in the patient. This is a significant limitation to the applicability of DL-based reconstruction methods in safety-critical applications. Dealing with data-dependent uncertainty and model uncertainty, also known as aleatoric (stochastic) and epistemic (systematic) uncertainty, respectively, has attracted much interest in computer vision over the last six years, along with effective tools for assessing the reliability of model predictions \cite{kendall2017,gal2016,abdar2021,hu2019,upadhyay2021,angermann2023eccv}. In the context of inverse problems, uncertainty quantification has been addressed in a Bayesian framework by sampling the posterior distribution \cite{bardsley2018computational,kaipio2006statistical}, which captures either epistemic or aleatory uncertainty alone. A framework for deep learning based Bayesian inversion has been established  in \cite{adler2018deep}.   Modeling uncertainty inherent in training data with respect to direct data consistent DL-based solvers requires further investigation.

In this work, we combine uncertainty estimation with data-consistent image reconstruction. For this purpose, we assume a parameterized Laplace distribution for the residuals between reconstructions and true images. Incorporating the uncertainty estimate into the reconstruction loss enables the simultaneous learning of a reconstruction and the corresponding uncertainty map. The uncertainty map can be used to infer the quality of a reconstructed image in the absence of ground truth. Data consistency is realized by null space networks. The approach is comprehensively tested using limited data problems in CT and MRI as practical use cases.

\section{Methods}
\label{sec:network_schemes}
	
Consider the inverse problem \eqref{eq:ip} where  $ \signal \in \R^n$ stands for the vectorized image  to be recovered and $\Ao \in \R^{m \times n} $ is the forward matrix.  In this section, we introduce the scheme of null space networks and other networks architectures that will be used as benchmarks. We discuss what adjustments can be made in the loss function during network training to simultaneously derive an uncertainty map while reconstructing the signal.
	
\subsection{Image reconstruction}

A reconstruction method for  \eqref{eq:ip}  is a family $(\Mo_\alpha)_{\alpha \in \Lambda}$ of potentially nonlinear operators   $\Mo_\alpha \colon  \R^m \to \R^n \colon \data \mapsto \Mo_\alpha (\data) $ mapping noisy data $\data \in \R^m$ to the reconstruction $\Mo_\alpha (\data) \in \R^n$. The crucial property is the convergence of the method  to a right inverse
	$\Mo_0 \colon  \ran(\Ao) \subseteq \R^m \to \R^n$  of the forward matrix  defined by  the property $\Ao \circ \Mo_0 \circ \Ao = \Ao$; see  \cite{nashed1987inner,Haltmeier2023}.
	Convergence of the reconstruction method means that for all $x \in \R^n$ we have
	\begin{equation} \label{eq:recon}	
		\sup_{\data} \norm{ \Mo_{\hat \alpha(\delta, \data)} (\data) - \Mo_0( \Ao \signal)}  \to 0 \text{ as } \delta \to 0\,,
	\end{equation}
	where the supremum is taken over all noisy data $\data \in \R^m$ with $\norm{\data - \Ao \signal} \leq \delta$ and the parameter $\alpha = \hat \alpha(\delta, \data) $ is chosen dependent on the noise level and the actual data. If  $\Lambda = \R_{>0}$ then $\hat \alpha$ is called parameter choice and   \eqref{eq:recon} means that $\left((\Mo_\alpha)_{\alpha >0}, \hat \alpha\right)$ is a regularization method for \eqref{eq:ip}, see \cite{engl1996regularization,scherzer2009variational}.
	
Classical reconstruction methods converge to  the Moore-Penrose inverse  $\Mo_0 = \Ao^\ddag$ of $\Ao$. If $\Ao$ has linearly independent columns, then $\Ao^\ddag \coloneqq (\Ao^* \Ao)^{-1} \Ao^*$ where $\Ao^*$ is the conjugate transpose.  Examples of such reconstruction methods  include Tikhonov regularization, truncated SVD or Landweber regularization. Different examples are variational regularization, where the right inverse $\Mo_0$ is defined as a quasi-solution having minimal value of a so-called regularization functional. More recent examples are neural network based methods  as we discuss next.

\subsection{Learned reconstruction}
\label{sec:networks}
	
In learned reconstruction methods, $(\Mo_\theta)_{\theta \in \Theta}$ consist of a rich family of mappings that typically depends on a high dimensional parameter  $\theta$ and includes neural networks. This allows  to extract  knowledge from available  training data such as databases of medical images~\cite{goodfellow2016}. NNs provide state-of-the-art for a variety of image processing tasks such as segmentation \cite{long2015,angermann2021}, image synthesis \cite{creswell2018,angermann2023eccv} and reconstruction \cite{gupta2018,yao2019}. The steady improvements in CNN architectures is evident in the increase in accuracy for medical imaging applications, with the U-net \cite{ronneberger2015} architecture still considered a high-performance model due to its ability to learn reliably in a limited data environment.
	
In the context of image reconstruction, popular approaches use networks of the form  \cite{jin2017,han2016deep}
	\begin{equation}\label{eq:twostep}	
		\Mo_{\alpha,\theta} \coloneqq    \No_\theta \circ \Bo_\alpha
	\end{equation}
	where $(\Bo_\alpha)_{\alpha > 0}$ is a classical reconstruction method and $(\Uo_{\theta})_{\theta \in \Theta}$ a neural network architecture. Here an initial reconstruction $\Bo_\alpha  (\data)$ is found based on data $\data$ and a network is trained to enhance this reconstruction according to the trained parameters. While $\Bo_\alpha$ serves as an approximate inversion of the forward operator, the learned part can be seen as a correction of the artifacts such aliasing artifacts for compressed sensing in MRI or streak artifacts in limited angle CT. Specifically, in our work we will use residual networks of the form
	\begin{align}
		\label{eq:resnet1}
		\Ro_{\theta}^{(1)}  &\coloneqq  \id+\Uo_{\theta_1},
		\\  \label{eq:resnet2}
		\Ro_{\theta}^{(2)}  &\coloneqq (\id+\Uo_{\theta_2} )  \circ (\id+\Uo_{\theta_1}),
	\end{align}
	defined by an arbitrary network architecture $(\Uo_{\theta})_{\theta \in \Theta}$, as reference method. In this study, $(\Uo_{\theta})_{\theta \in \Theta}$ is considered a basic U-net architecture unless stated otherwise.

\subsection{Data consistency}
\label{sec:null-space}
	
Given an initial reconstruction $ \Bo_\alpha (\data)$ and any network $\Uo_{\theta}$, the two step method  \eqref{eq:twostep} does not promote data-consistent solutions. This means that even  if $\Ao (\Bo_\alpha (\data)) $ is close to the data $\data$ this is not necessarily the case  for $ \Ao (\Uo_{\theta} \circ \Bo_\alpha)( \data) $. To address this issue, one approach is to consider variational or iterative networks \cite{hammernik2018learning,adler2017solving,yiasemis2022,genzel2022,schlemper2017deep,kofler2018u}. However, models with an iterative architecture increase training time and controlling the data discrepancy  $\norm{ \Ao \signal - \data}^2$ is still challenging.

In order to derive a data-consistent solution from any reconstruction $\signal_\Mo \coloneqq \Mo( \data)$  we therefore consider the orthogonal projection 
\begin{equation} \label{eq:PL}
 \Po_\solspace (\signal_\Mo) = \argmin_\signal \{ \lVert \signal^\Mo - \signal \rVert \mid \Ao \signal = \data \}
\end{equation}
onto the solution space $\solspace(\Ao,\data)  \coloneqq \{\signal \mid \Ao \signal = \data \}$. An iterative way to calculate the orthogonal projection is given by the Landweber iteration  \cite{engl1996,landweber1951iteration}
\begin{equation}\label{eq:landweber}
	\forall j \in \N \colon \quad 
	\signal_{j+1} = \signal_j - \lambda_j \Ao^* (\Ao \signal_j  - \data) \,,
\end{equation}
with initial value  $\signal_0 =  \signal_\Mo$, where $\lambda_j>0$ is the stepsize for the $j$-th iteration. Note that the Landweber iterations \eqref{eq:landweber} will be applied to the output of an already trained network. This allows for a large number of iterations, which ensures a good approximation of the projection onto $\solspace$ and does not prolong the training process. In the ill-posed case, regularization can be integrated into \eqref{eq:landweber} by early stopping,  by replacing  the forward model $\Ao$ in with an operator having a stable  pseudoinverse, or by replacing 
the data $\data$ with  $(\Ao \circ \Bo_\alpha) (\data)$ in a two-step method $ \Mo_{\alpha, \theta} = \No_\theta \circ \Bo_\alpha $.

\subsection{Null space networks}
	
\begin{figure}[htb!]
		\centering
		\begin{tikzpicture}[scale=.89]
			\node at (-1, 0) {$x$};
			\draw[line width=1.3pt,-stealth] (-0.75, 0)--(0, 0);
			\draw[draw=black,fill=gray] (0,-2) rectangle ++(0.25, 4);
			\draw[line width=1.3pt,-stealth] (0.3, 0)--(1, 0);
			\draw[draw=black,fill=gray] (1,-2) rectangle ++(0.25, 4);
			\draw[line width=1.3pt,-stealth] (1.3, 0)--(2, 0);
			\draw[draw=black,fill=gray] (2,-2) rectangle ++(0.25, 4);
			\draw[line width=1.3pt,-stealth] (2.3, 0)--(3, 0);
			\node at (3.5, 0) {$\ldots$};
			\draw[line width=1.3pt,-stealth] (4, 0)--(4.7, 0);
			\draw[draw=black,fill=gray] (4.7,-2) rectangle ++(0.25, 4);
			\draw[line width=1.3pt,-stealth] (5, 0)--(5.7, 0);
			\draw[draw=black,fill=blue] (5.7,-2) rectangle ++(0.25, 4);
			\node at (6.35, 0) {$+$};
			\node at (7.5, 0) {$x = \Psi_{\theta}(x)$};
			
			\draw[->] (-1, 0.25)  to [out=90,in=90, looseness=1.5] (6.7, 0.25);
			\node at (3, 4) {$\id$};
			\node at (6, -2.5) {${\color{blue} \Po_0}$};
			
			\draw [decorate,decoration={brace,amplitude=5pt,mirror,raise=4ex}]
			(0,-1.75) -- (5,-1.75) node[midway,yshift=-3em]{$\Uo_\theta$};
		\end{tikzpicture}
		\caption{\textbf{Null space network.} \label{fig:null} Illustration of a null space network $\id + \Po_{0} \Uo_{\theta}$. The last layer projects the output of the residual part $\Uo_{\theta}$ on the null space of forward operator $\Ao$.}
\end{figure}
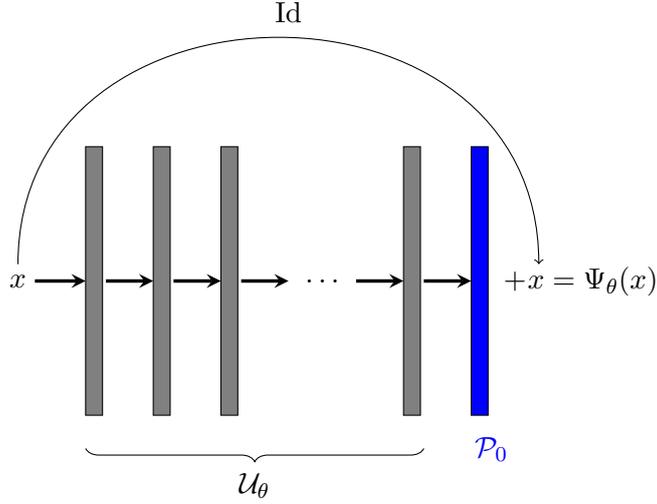

In order to obtain solutions that are guaranteed to be data consistent,  the concept of null space networks has been introduced in \cite{schwab2019}. Null space networks basically denote residual networks that only modify components in the kernel of the forward  problem.  More formally, let 
$(\Uo_{\theta})_{\theta \in \Theta}$ be any network architecture and $\Po_0$ denote  the orthogonal projection onto the null space of $\Ao$. We then call
\begin{align} \label{eq:nullnet1}
\Psi^{(1)}_{\theta}  &\coloneqq  ( \id + \Po_0 \circ \Uo_{\theta} )\,,
	\\  \label{eq:nullnet2}
\Psi^{(2)}_{\theta}  &\coloneqq  ( \id + \Po_0 \circ \Uo_{\theta_2} ) \circ ( \id + \Po_0 \circ \Uo_{\theta_1} ) ,
\end{align}
null space network associated to operator $\Ao$ with architecture  $(\Uo_{\theta})_{\theta \in \Theta}$ and cascade length one and two, respectively. An illustration of a null space network (with cascade length one) is shown in Figure \ref{fig:null}.

We use null space networks in the context of  the two-step reconstruction \eqref{eq:twostep}.   Note  that any  null space network  is a particular form  of a residual network, where the residual correction only operates in the null space of the forward operator $\Ao$ and by that ensuring data consistency $\Ao \circ \Psi^{(2)}_{\theta}  = \Ao \circ \Psi^{(1)}_{\theta} = \Ao$.  One readily verifies that $  (  \id + \Po_0 \circ \Uo_{\theta}  ) ( \Ao^\ddag \data)
=   \Po_\solspace ( \id +   \Uo_{\theta} ) ( \Ao^\ddag \data) $. Therefore  two-step reconstructions using either a null space network $\id + \Po_0 \circ \Uo_{\theta}$  or  a projected residual network $\Po_\solspace  \circ ( \id +  \Uo_{\theta})$  for given  parameters coincide   \cite{antholzer2018deep}. However, training a null-space network  is clearly different  from training the residual network followed by a projection onto $\Po_\solspace$.           
	
\subsection{Network training}
\label{sec:training}
	
Let  $(\Mo_\theta)_{\theta \in \Theta} $  be a reconstruction method whose parameter vector $\theta$  is selected  based on training data  $(\Ao \signal_i, \signal_i)_{i=1}^N$. Note that the subscript $\alpha$ for the classical reconstruction $(\Bo_\alpha)_{\alpha > 0}$ in \eqref{eq:twostep} is omitted here for simplicity. The common training procedure minimizes the empirical risk
\begin{equation} \label{eq:risk}
\risk (\theta) \coloneqq \frac{1}{N} \sum_{i=1}^{N} \norm{\Mo_\theta(\Ao \signal_i) - \signal_i }_1
\end{equation}
According to  \cite{upadhyay2021,angermann2023eccv}, the underlying assumption can be interpreted that every component of the residual image $\left(\epsilon_{\theta,p}\right)_{p=1}^n \coloneqq  \Mo_\theta(\Ao x) - x$ follows a Laplace distribution $\epsilon_{\theta,p} \sim \operatorname{Laplace}(0,\sigma_{})$ with density $\exp\left(- \abs{\epsilon_{\theta,p}}/\sigma_{}\right)/(2\sigma_{}).$ Maximum likelihood optimization  yields
	\begin{align*}
		\hat \theta& =
		\argmax_\theta \frac{1}{N} \prod_{i,p} \frac{1}{2\sigma}
		\exp\left(-\abs{ \epsilon_{\theta,p} }  / \sigma_{}\right)
		\\
		&=
		\argmin_\theta \frac{1}{N} \sum_{i,p} \lvert (\Mo_\theta(\Ao \signal_i) - x_{i})_p \rvert /\sigma_{} + \frac{1}{N} \log(2\sigma_{})
		\\
		&=
		\argmin_\theta \frac{1}{N} \sum_{i} \norm{ \Mo_\theta(\Ao \signal_i)- \signal_{i} }_1 /\sigma_{}
		+ \frac{1}{N} \log(2\sigma_{}) \,.
	\end{align*}
Hence, for pixel independent scale parameter $\sigma$, maximum likelihood optimization     minimization recovers minimization of the empirical risk~\eqref{eq:risk}.

\subsection{Uncertainty quantification}
\label{sec:uncertainty}
	
Obviously, the assumption of a fixed scale for all residuals $\epsilon_{\theta,p} \coloneqq  (\Mo_\theta(\Ao x) - x)_p$ is quite strong.  In the context of uncertainty quantification it can be relaxed  considering input-dependent scale parameter. To do so, the scale is now taken as  a whole image $\sigma = (\sigma_p)_{p=1}^n$ modeled as a function of data $\data \in \R^m$.   More precisely, the network now simultaneously predicts a reconstruction $\signal_\theta = \Mo_\theta^\signal(\data)$ and its corresponding scale map $\sigma_\theta = \Mo_\theta^\sigma(\data)$. In our implementation, the reconstruction network $\Mo_\theta^\signal$ will be taken as a two-step architecture \eqref{eq:twostep} combined with the  null space network \eqref{eq:nullnet1} or \eqref{eq:nullnet2}. The uncertainty estimating network  $\Mo_\theta^\sigma$ has the similar  architecture and nearly shares all parameters. More precisely, we split the network architecture $\Uo_\theta$  in the last null space block to obtain two output-branches, where one gives the data-consistent reconstruction and one the uncertainty map. For further implementation details we refer to the available github repository \url{https://github.com/anger-man/cascaded-null-space-learning}.

The networks $\Mo_\theta^\signal$ and $\Mo_\theta^\sigma$ are trained by simultaneously minimizing the  uncertainty-aware  loss
\begin{equation} \label{eq:riskns}
	\riskunc (\theta)\coloneqq\frac{1}{N} \sum_{i,p} \frac{\abs{(\Mo_\theta^\signal(\Ao \signal_i)-\signal_i)_p}}{(\Mo_\theta^\sigma(\Ao \signal_i))_p}
	+ \frac{1}{N}\sum_{i,p} \log\left(2\cdot (\Mo_\theta^\sigma(\Ao \signal_i))_p\right) \,.
	\end{equation}
By taking $(\Mo_\theta^\sigma(\Ao \signal_i))_p = \sigma$ as a constant,  \eqref{eq:riskns} would reduce to  \eqref{eq:risk}. However simultaneous minimization allows  both reconstructing the image and the associated uncertainty. For regions with large absolute residuals we  obtain large values in the scale map corresponding to high uncertainty. At the same time, the logarithmic term penalizes the model to avoid predicting high uncertainty for all pixel regions.
	
The benefits of uncertainty-aware models are versatile. The training process is more robust against OOD data in the training set, insufficient measurement quality is established during prediction, and unwanted objects or artifacts that may effect the reconstruction are detected and localized in the absence of ground truth data.

	\subsection{Implemented methods}
	
	In our simulations  we compare a total of eight different learned reconstruction methods  against the ground truth and the Moore-Penrose inverse:
	\begin{itemize}
		\itemsep .5em
		\item $\signal_0$: Ground truth used for error evaluation.
		
		\item$\signal^\ddag=  \Ao^\ddag (\data)$: Pseudoinverse reconstruction.
		
		\item $\signal_{\Ro,1} = \Ro_{\theta}^{(1)} (\signal^\ddag)$: Two-step reconstruction \eqref{eq:twostep} +  \eqref{eq:resnet1}.
		
		\item $\signal_{\Ro,2} = \Ro_{\theta}^{(2)} (\signal^\ddag)$: Two-step reconstruction \eqref{eq:twostep} + \eqref{eq:resnet2}.
		
		\item $\Po_{\solspace} \left(\signal_{\Ro,1}\right)$: Projection of $\signal_{\Ro,1}$ onto $\solspace(\Ao,\data)$ \eqref{eq:PL}.
		
		\item $\Po_{\solspace} \left(\signal_{\Ro,2}\right)$: Projection of $\signal_{\Ro,2}$ onto $\solspace(\Ao,\data)$ \eqref{eq:PL}.
		
		\item $\signal_{\Psi,1} = \Psi_{\theta}^{(1)} (\signal^\ddag)$: Null space   reconstruction \eqref{eq:twostep} +  \eqref{eq:nullnet1}.
		
		\item $\signal_{\Psi,2} = \Psi_{\theta}^{(2)} (\signal^\ddag)$: Null space   reconstruction \eqref{eq:twostep} +  \eqref{eq:nullnet2}.
		
		\item $\signal_{\Psi,1}^\text{unc} $: Uncertainty-aware reconstruction using \eqref{eq:nullnet1}.
		
		\item $\signal_{\Psi,2}^\text{unc} $: Uncertainty-aware reconstruction using \eqref{eq:nullnet2}.
		
	\end{itemize}
	
	The underlying networks $\Ro_{\theta}^{(1)}$, $\Ro_{\theta}^{(2)}$, $\Psi_{\theta}^{(1)}$, $\Psi_{\theta}^{(2)}$  are trained using the MAE loss function~\eqref{eq:risk}, except for the uncertainty-aware networks $\signal_{\Psi,1}^\text{unc} $, $\signal_{\Psi,2}^\text{unc} $, which are trained  by minimizing the uncertainty aware loss~\eqref{eq:riskns}. For the reconstructions  $\Po_{\solspace} \left(\signal_{\Ro,1}\right)$ and  $\Po_{\solspace} \left(\signal_{\Ro,2}\right)$ the projection onto the space  of all solutions of $\Ao \signal = \data$  is calculated  using 15 iterations of the Landweber iteration. Reconstruction quality is assessed by peak-signal-to-noise ratio ($\psnr$) and the structural-similarity index (SSIM).

\section{Results}
\label{sec:results}
	
In this section we present the numerical results for the discussed reconstruction methods. To this end, we will consider two examples of inverse problem of the form \eqref{eq:ip}: (A) Reconstruction from subsampled Fourier measurements on fastMRI dataset \cite{zbontar2018,knoll2020}; (B) reconstruction of limited angle Radon measurements on a self-acquired toy dataset. Note that it is not the aim of this study to propose a new state-of-the-art model to solve specific  inverse problems. With these experiments, we show that simple adjustments to the network architecture and slight modifications to the loss function result in a significant increase in reconstruction quality and the benefit of pixel-based confidence prediction for each reconstruction. Note also that these modifications are implemented and evaluated considering the basic U-network architecture widely used in image processing tasks. However, our rationale can be applied to any network architecture.

\subsection{Study A: 4-fold accelerated MRI}

The first case study is devoted to a real-world scenario of compressed sensing in MRI. To this end, we use the public fastMRI dataset, which consists of 1594 multi-coil knee MRI scans (\url{https://fastmri.med.nyu.edu/}). The experiments are based on the subset of 571 randomly selected diagnostic cases without fat-suppression. For a practice-oriented evaluation on unseen individuals, \SI{5}{\percent} of the scans have been randomly selected to represent the test cases. This yields nearly $\si{20000}$ samples for training and $\si{1000}$  samples for testing. For the training set $(\signal_i)_{i=1}^N$, we draw magnitude images $\signal_i \in \R^{320\times 320}$ obtained by fully sampled multi-coil data. Similar to \cite{genzel2022}, our model corresponds to the simpler modality of single-coil MRI. The fastMRI challenge also provides emulated single-coil data,  that is drawn in a retrospective way from the multi-coil measurements. However, we decided to sample from the multi-coil reconstructions in the favor of higher image quality and noise-reduced measurements.

	\begin{figure*}[htb!]
		\centering
		\subfloat[$\signal_0$]{\includegraphics[width=0.32\textwidth]{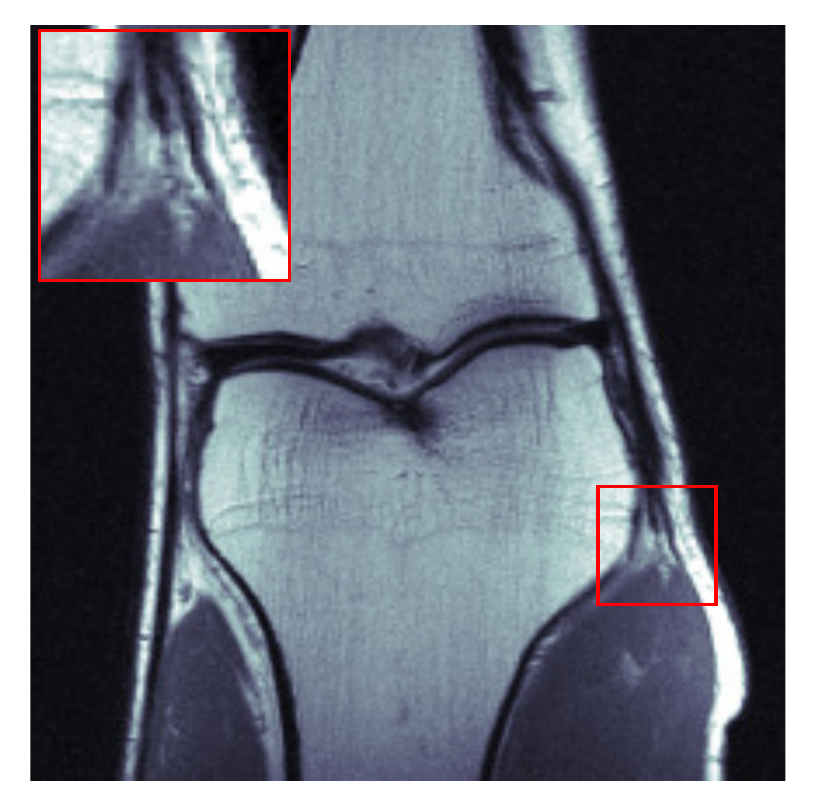}
			\label{fig:ground truth}}
		\subfloat[$\signal^\ddag$]{\includegraphics[width=0.32\textwidth]{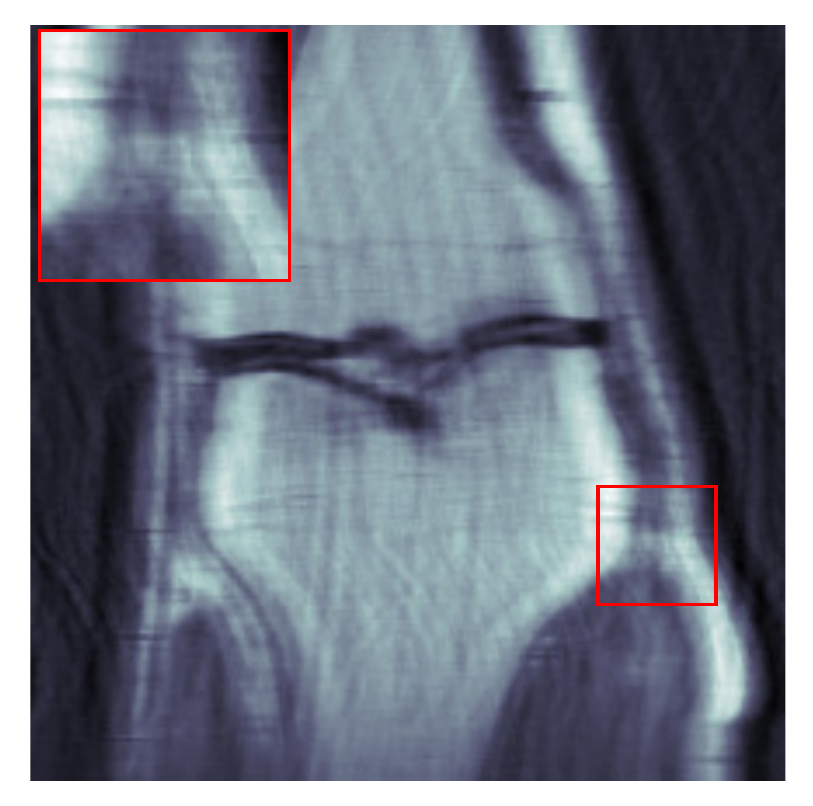}
			\label{fig:pseudo inverse}}
		\subfloat[$\Po_\solspace(\signal_{\Ro,1})$]{\includegraphics[width=0.32\textwidth]{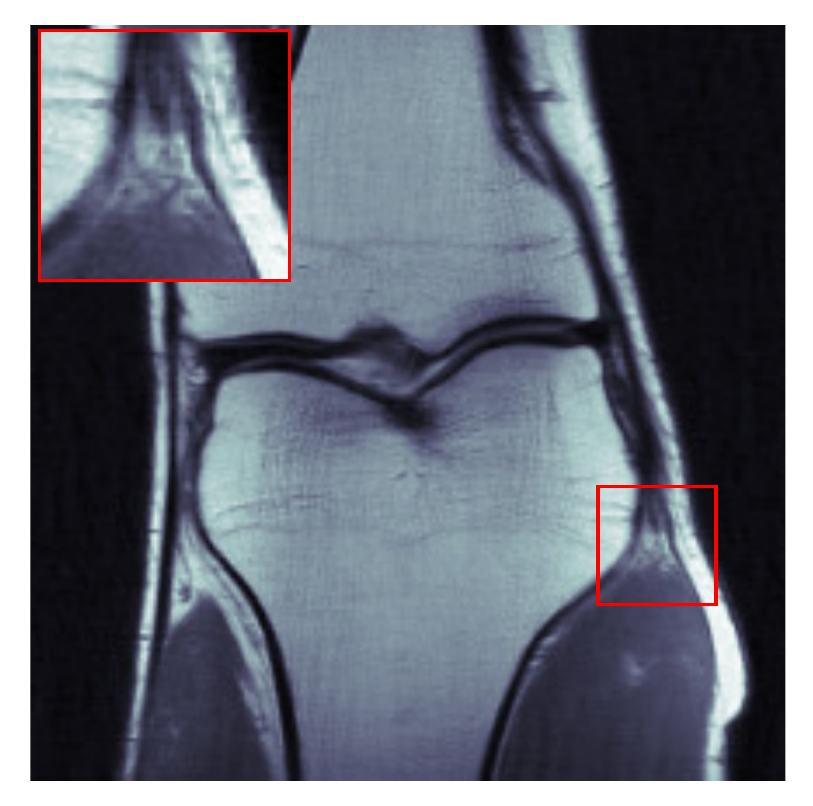}
			\label{fig:cascaded resnet}}\\ \vspace{-.5em}
		\subfloat[$\Po_{\solspace} \paren{\signal_{\Ro,2}}$]{\includegraphics[width=0.32\textwidth]{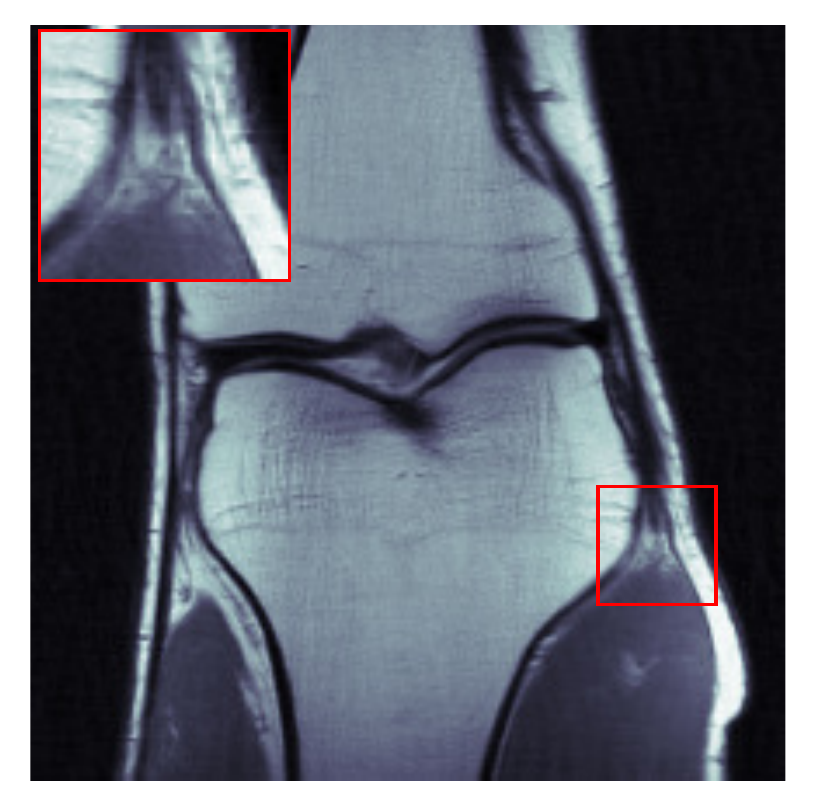}
			\label{fig:iter cascaded unet}}
		\subfloat[$\signal_{\Psi,1}$]{\includegraphics[width=0.32\textwidth]{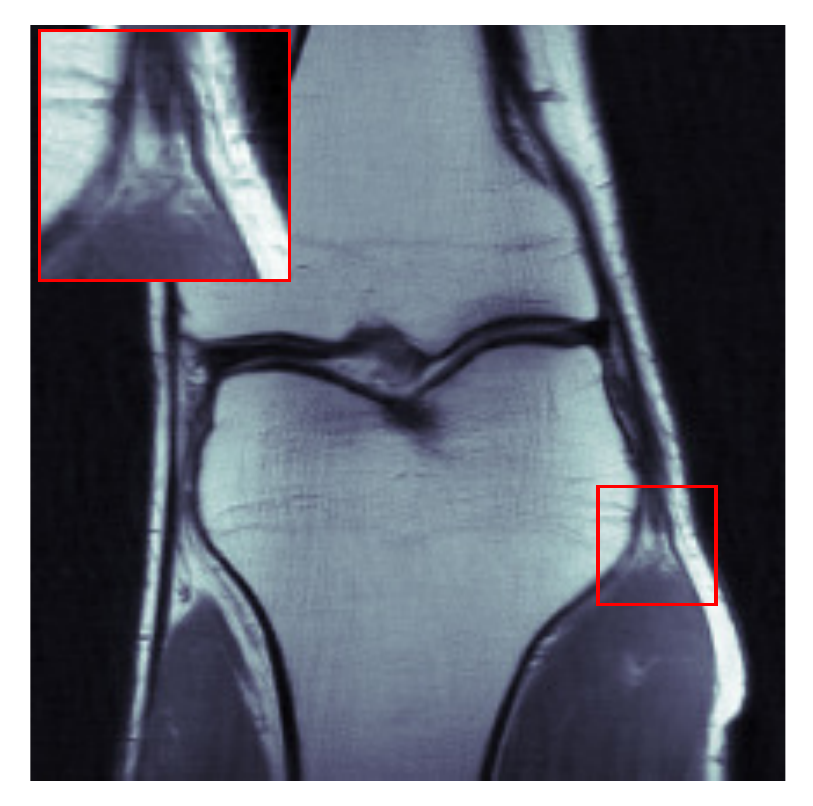}
			\label{fig:null space}}
		\subfloat[$\signal_{\Psi,2}$]{\includegraphics[width=0.32\textwidth]{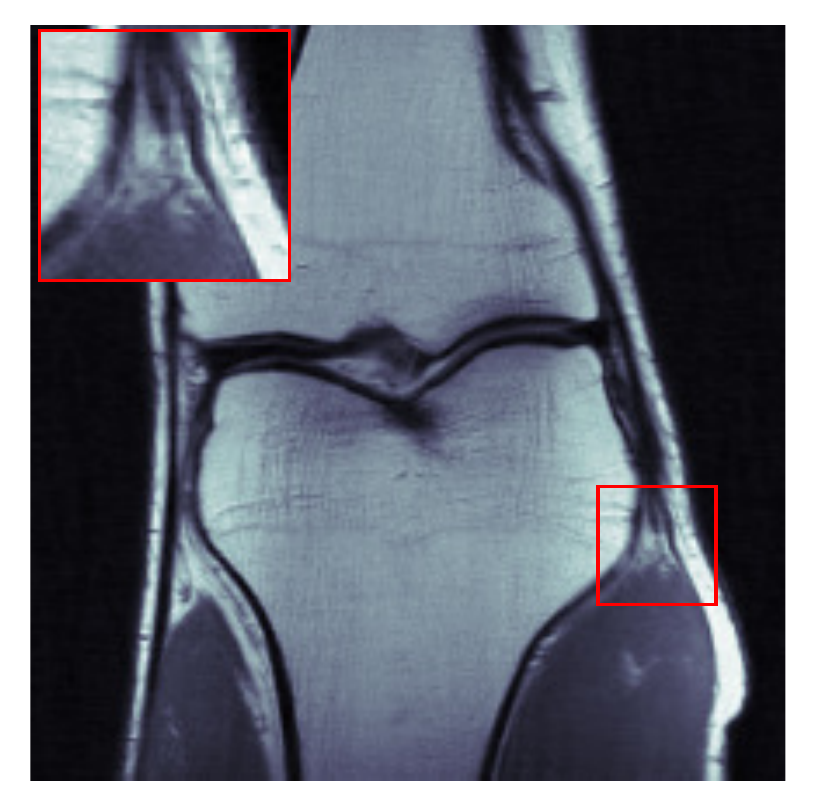}
			\label{fig:cascaded null space}}
		\caption{\textbf{Accelerated MRI reconstruction results.} Ground truth image $x_0$  (central slice of a randomly selected volume from the test set),  pseudoinverse reconstruction $\signal^\ddag$, projected two-step reconstructions $\Po_{\solspace} \paren{\signal_{\Ro,1}}$, $\Po_{\solspace} \paren{\signal_{\Ro,2}}$ and null-space reconstructions $\signal_{\Psi,1}$, $\signal_{\Psi,2}$.}
		\label{fig:mri}
	\end{figure*}	
	
\subsubsection*{Implementation details}
	
Here, the  forward operator takes  the form $\Ao = \So \circ \Fo$, where $\Fo \colon \C^{320 \times 320} \to \C^{320\times 320}$   is the two-dimensional discrete Fourier transform and $\So \colon \C^{320\times 320}  \to \C^{320\times 320}$ is a subsampling operator, implemented via a binary mask omitting Fourier measurement lines in the phase-encoding direction. The amount of dropped k-space lines depends on the acceleration factor and in our case is set to \SI{75}{\percent}, following the subsampling scheme in \cite{hyun2018}. The Fourier transform is a unitary operator. Therefore, the Moore-Penrose inverse of $\Ao$  is given by the conjugate transpose of the forward operator, $\Ao^\ddag = \Ao^* = \Fo^* \circ \So$. Note that we actually have complex-valued data. Thus before feeding the instances to a neural network function, we concatenate real and imaginary part such that they are treated as separat channels,  $\signal^\ddag \in \R^{2(320\times 320)}$. The data processing pipeline and the models are implemented in Python using PyTorch library (\url{https://pytorch.org/}) for GPU-accelerated computation. For any details on training and hyperparameter selection, we refer to our github repository \url{https://github.com/anger-man/cascaded-null-space-learning}.

	\begin{table}[htb!]
		\normalsize
		\begin{center}
			\caption{\textbf{Accelerated MRI.} Quantitative evaluation on nearly \si{1000} test slices. The reported metrics are $\psnr$ and $100 \times \ssim$  (higher is better).}
			\begin{tabular}{l | c | c | c | c }
				\toprule
				\multicolumn{5}{c}{\textbf{benchmark}}\\ \midrule
				& $\signal_{\Ro,1}$ & $\Po_\solspace(\signal_{\Ro,1})$ & $\signal_{\Ro,2}$  &$\Po_\solspace(\signal_{\Ro,2})$\\ \midrule
				$\psnr$ &31.29&32.12&31.5&32.35 \\ \midrule
				$100 \times \ssim$ &85.13&86.17&85.44&86.5 \\ \midrule
				\multicolumn{5}{c}{\textbf{null space networks}}\\ \midrule
				&$\signal_{\Psi,1}$& $\signal_{\Psi,1}^\text{unc}$ 	&$\signal_{\Psi,2}$& $\signal_{\Psi,2}^\text{unc}$\\ \midrule
				$\psnr$ &32.25&32.28&33.39&\textbf{33.44 }\\ \midrule
				$100 \times \ssim$ &86.31&86.4&88.2&{{\textbf{88.29}}} \\ \bottomrule
			\end{tabular}
			\label{tab:mri}
		\end{center}
	\end{table}

\subsubsection*{Reconstruction results}
	
The quantitative results shown in Table~\ref{tab:mri} demonstrate the superior accuracy of DL based solvers. The benchmark $\signal_{\Ro,1}$ already achieves  a PSNR of 31.29 and a SSIM of 0.851 for the set of nearly 1k unseen test slices. The result is quite impressive when we consider an acceleration rate of 4 in the compressed sensing scenario, i.e., only \SI{25}{\percent} of the lines in phase encoding directions have been kept.  The projection of the previously trained basic method onto the solution space $\solspace$ leads to approximately data-consistent solutions. Its impact is confirmed by a slight increase for both test metrics. The extension of $\signal_{\Psi,1}$ to a cascaded scheme $\signal_{\Psi,2}$ has a positive impact on the entire optimization procedure as indicated by a significant increase to 33.39 and 0.882 for PSNR and SSIM metric, respectively.
	
In Figure~\ref{fig:mri} we observe the prominent aliasing artifacts in the pseudoinverse  reconstruction $\signal^\ddag = \Ao^\ddag (\data)$. These artifacts are obviously removed in all learned reconstructions. The examination of the magnified areas shows that the contrast in the high-frequency details is higher for the null space networks  than for the residual networks. Using the  null space approach for joint reconstruction and uncertainty prediction does not significantly change the reconstruction quality. Therefore, it is possible to incorporate uncertainty awareness into the cascade of null space blocks without affecting the recovery performance at all. Benefits of an uncertainty estimate are investigated in the following.

\begin{figure}[htb!]
\centering
\includegraphics[width=0.32\textwidth]{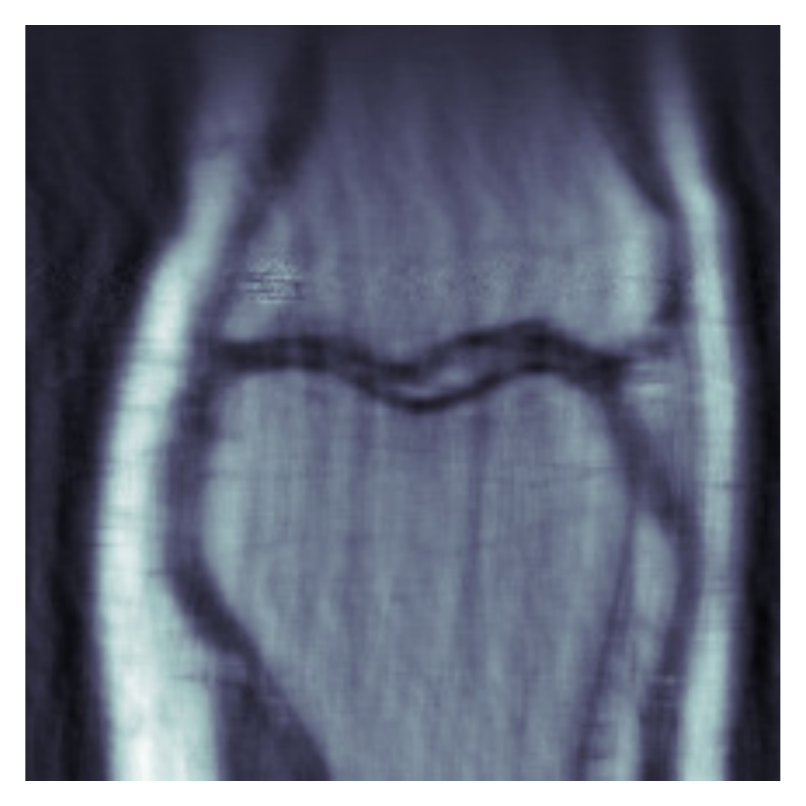}
\includegraphics[width=0.32\textwidth]{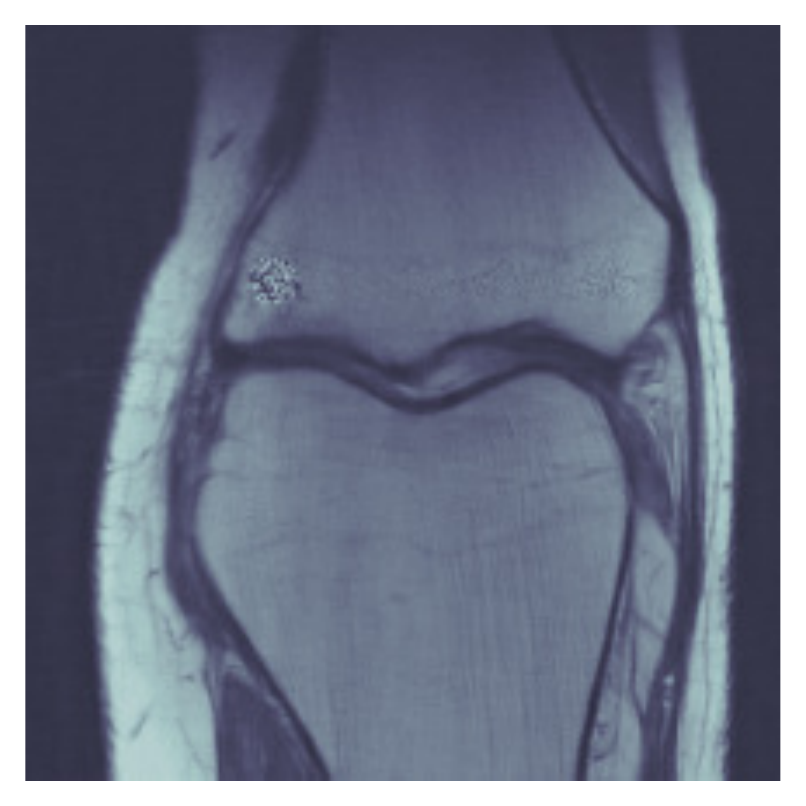}
\includegraphics[width=0.32\textwidth]{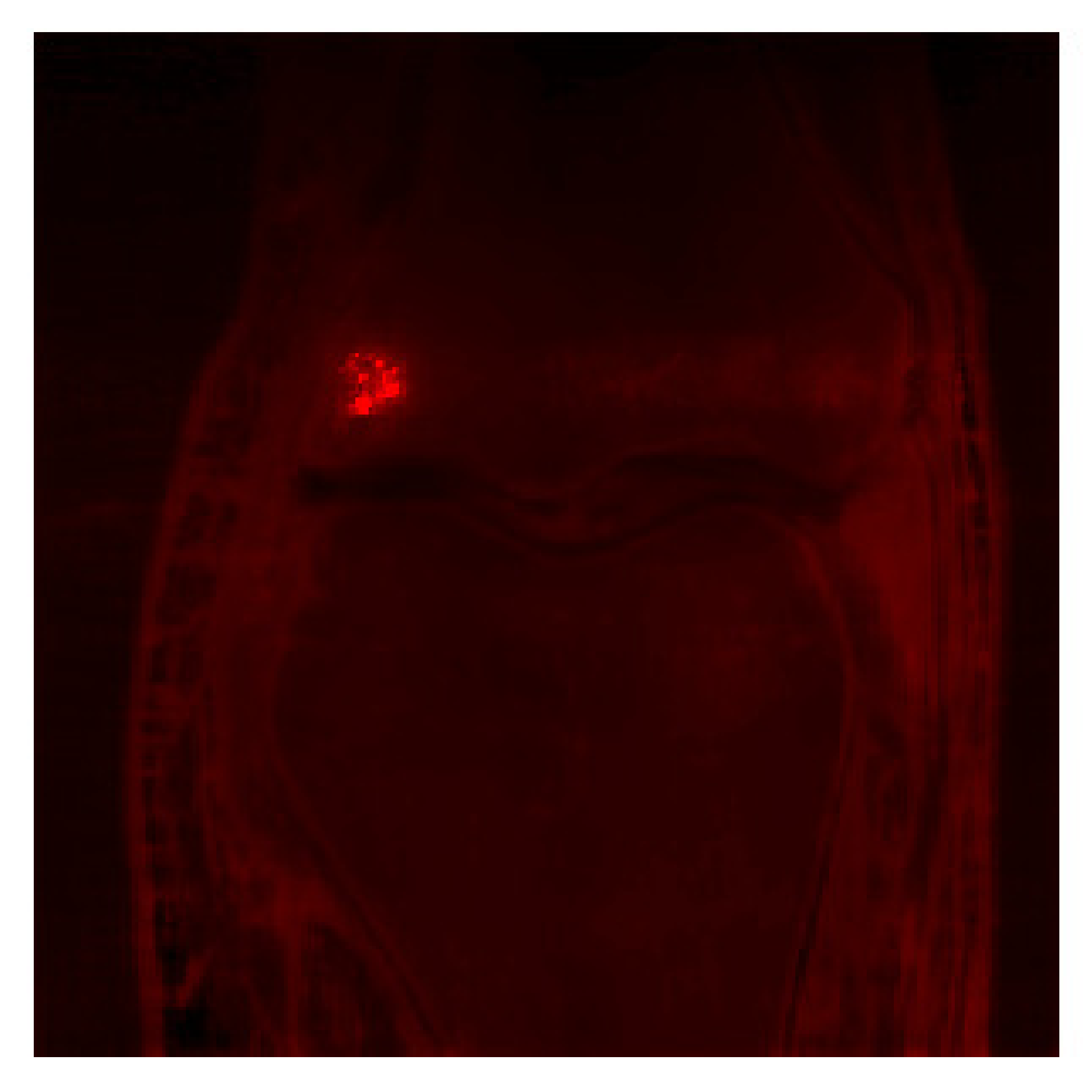}
\\
\includegraphics[width=0.32\textwidth]{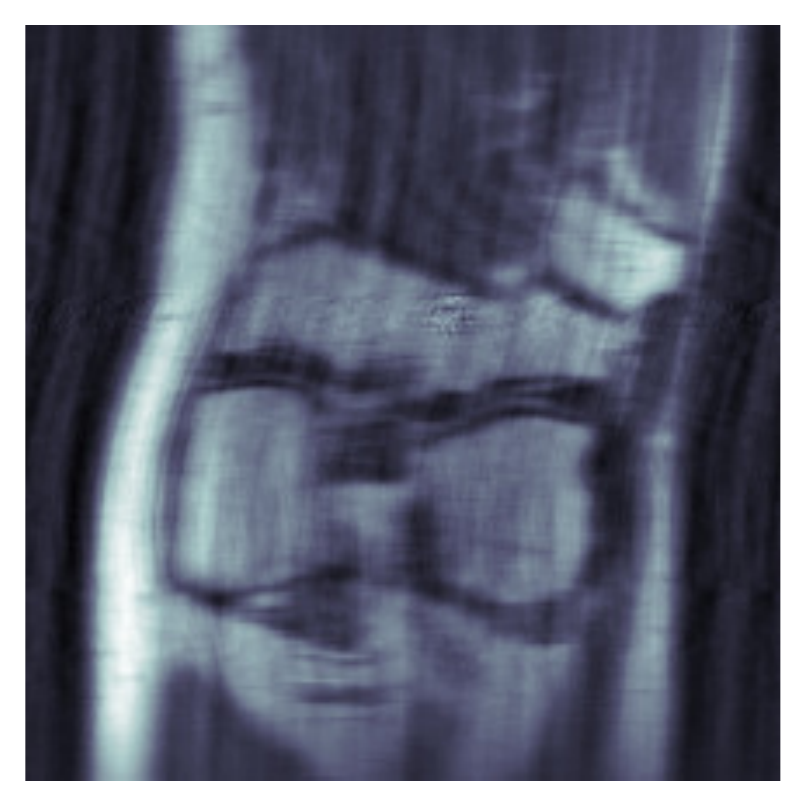}
\includegraphics[width=0.32\textwidth]{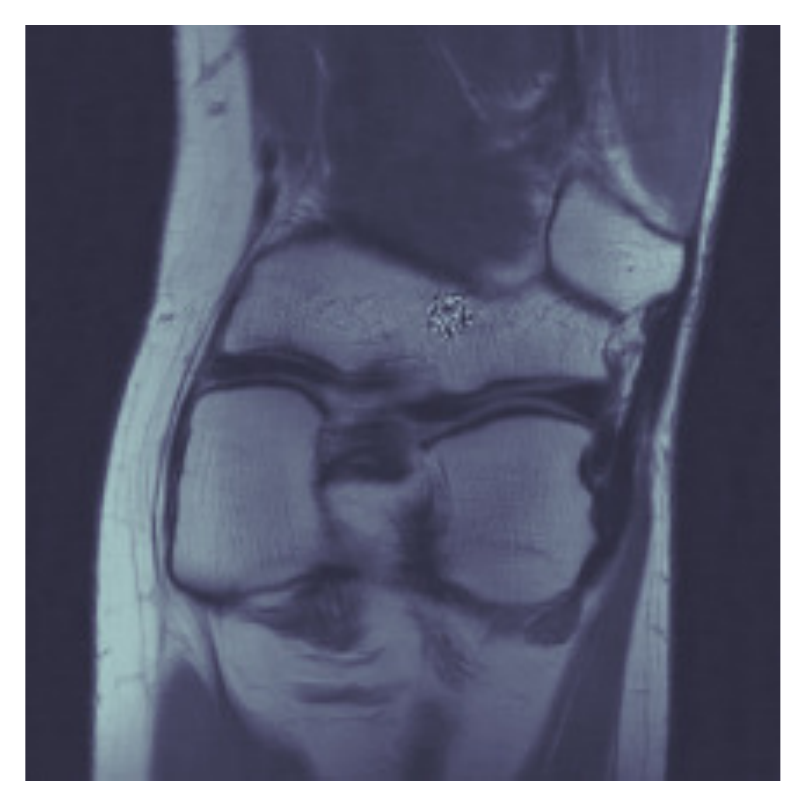}
\includegraphics[width=0.32\textwidth]{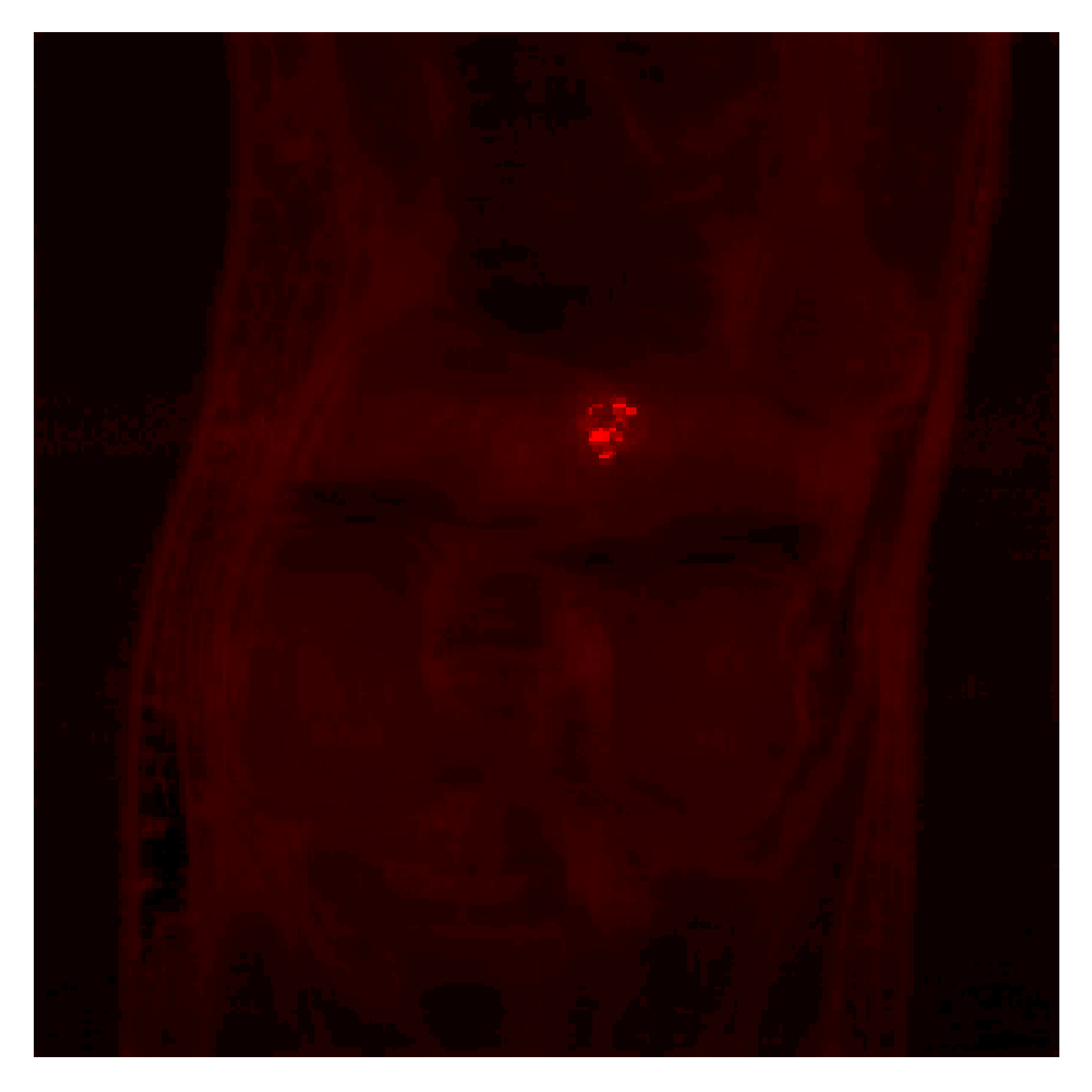}
\caption{\textbf{Position-related uncertainty in  accelerated MRI}.  The presence of an atypical irregularity in the knee causes artifacts on the entire horizontal trace of the pseudo-inverse solution $\signal^\ddag$ (first column) and leads to an unpredictable behavior of the reconstruction $\signal_{\Psi,2}^{\rm unc} $ at the corresponding locations (second column). In the simultaneously predicted uncertainty maps (third column), these irregular objects are clearly indicated by higher uncertainty values.}
\label{fig:fastmri_unc}
\end{figure}

\subsubsection*{Uncertainty quantification}

The uncertainty reconstruction provides a pixel-wise confidence value additionally to the reconstruction, which can be used to detect suspicious objects that are not covered by the training distribution. Furthermore, uncertainty evaluation based on these region-based confidence maps can serve as a quality assessment of the overall prediction. In the following, both advantageous properties are investigated numerically.

Unknown local properties of a patient's knee under investigation can lead to unpredictable behavior of the reconstruction method in that region. Unknown properties here refer to the occurrence of non-ubiquitous objects such as cancer or bone fractures. We assume here that the presence of these objects is true, i.e., the irregularity is not caused by the measurement process itself.  For the simulations, we select two magnitude images from the test cases and add salt-and-pepper noise to small subregions. We then apply the forward operator to obtain measurement data $\data_\text{cor}$ describing the actual measured data if such an irregularity would be present in the examined knee. The results of applying the Moore-Penrose inverse $\Ao^\ddag$ and the uncertainty aware null space approach to $\data_\text{cor}$  are shown in Figure~\ref{fig:fastmri_unc}.  As can be seen, the predicted uncertainty maps show large values at the corresponding location, visualized by a brighter color intensity. We claim that this characteristic of the uncertainty-aware null space networks clearly identifies unreliable regions in the reconstruction caused by the presence of abnormal features in an examined knee.

\begin{figure}[htb!]
\centering
\includegraphics[width=.7\columnwidth]{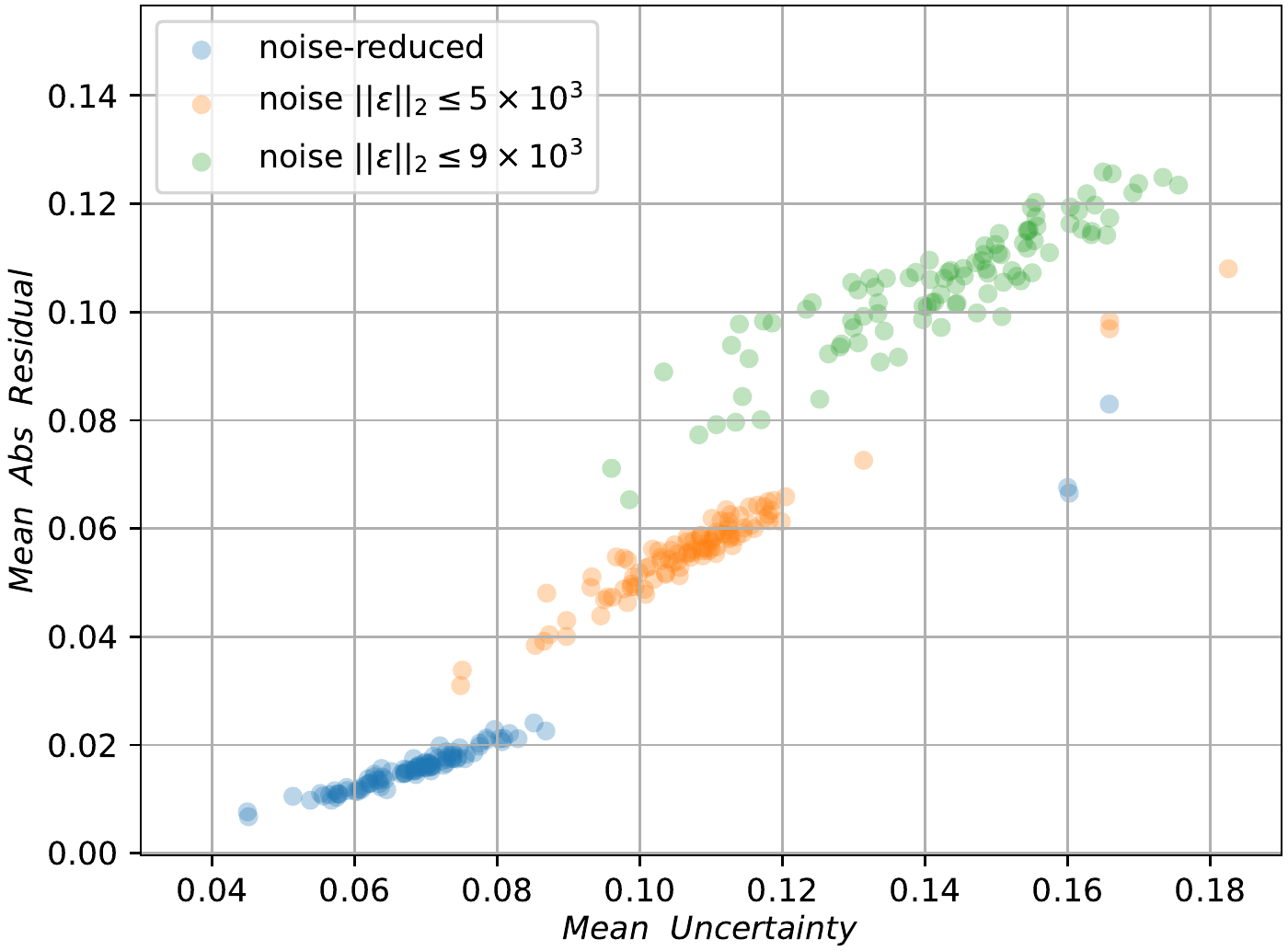}
\caption{\textbf{Prediction error versus mean uncertainty for accelerated MRI.} A clear correlation is found between the prediction error and the average value of the modeled uncertainty maps. Perturbations of the measurement data simultaneously  increases prediction error  and  mean uncertainty score.}
\label{fig:fast_quant}
\end{figure}

The experiments in this study  are based on the root-sum-of-squares reconstruction of multi-coil measurements. As a consequence, the slices in the training set $(\signal_i)_{i=1}^N$ denote noise-reduced samples. We select 100 test slices and perturb the measured data by additive noise, i.e. we generate measurements $\data_i = \Ao \signal_i+\epsilon$ for $i=1,\ldots,N$, where $\norm{\epsilon}_2\leq \delta$ for $\delta \in  \{0,\num{5e3},\num{9e3}\}$. For each test slice, we define the average pixel value of the uncertainty map as the image-based uncertainty score and plot it against  the corresponding mean absolute residual between reconstruction and noise-free ground truth in Figure~\ref{fig:fast_quant}. In the noise-reduced case $\epsilon = 0$, we infer a fairly clear positive correlation between the mean uncertainty and the mean absolute residuals. For $\norm{\epsilon}_2\leq \num{5e3}$ the prediction error as well as the  uncertainty values increase. The effect is amplified by increasing the scale of the additive Gaussian noise to $\norm{\epsilon}_2\leq \num{9e3}$.  The perturbation of the measured data during testing leads to OOD data, which is clearly indicated by an increase in uncertainty scores. We conclude that the proposed method can be used to assess reconstruction quality (correlation with MAE) and to detect out-of-distribution data (larger scale for noise-perturbed data).

\begin{figure}[htb!]
\centering
\subfloat[$\signal_0$]{\includegraphics[width=0.3\textwidth]{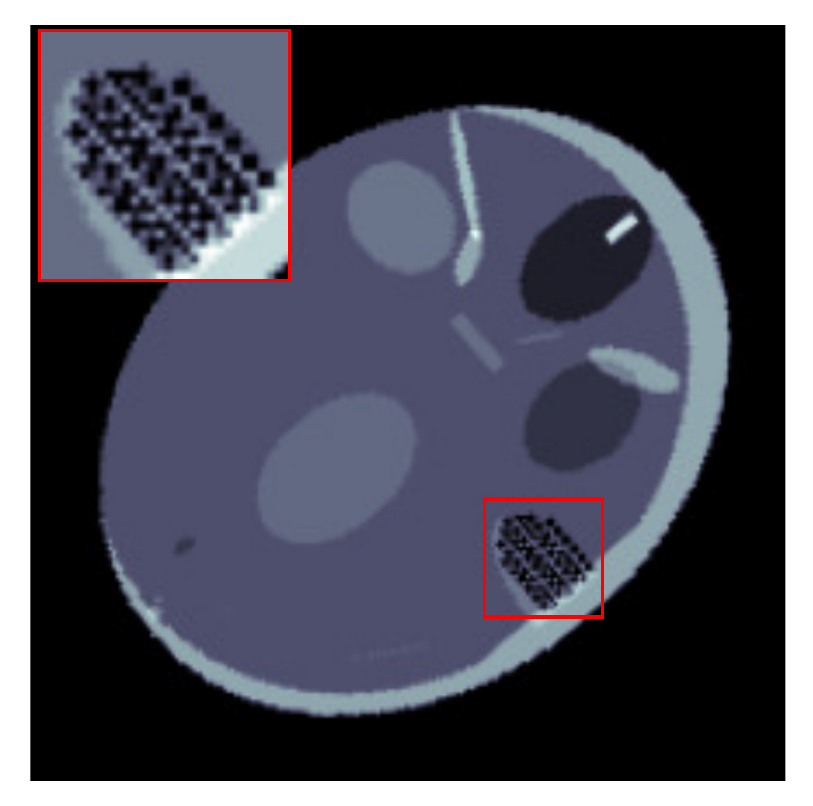}
\label{fig:CT ground truth}}
\subfloat[$\signal^\ddag$]{\includegraphics[width=0.3\textwidth]{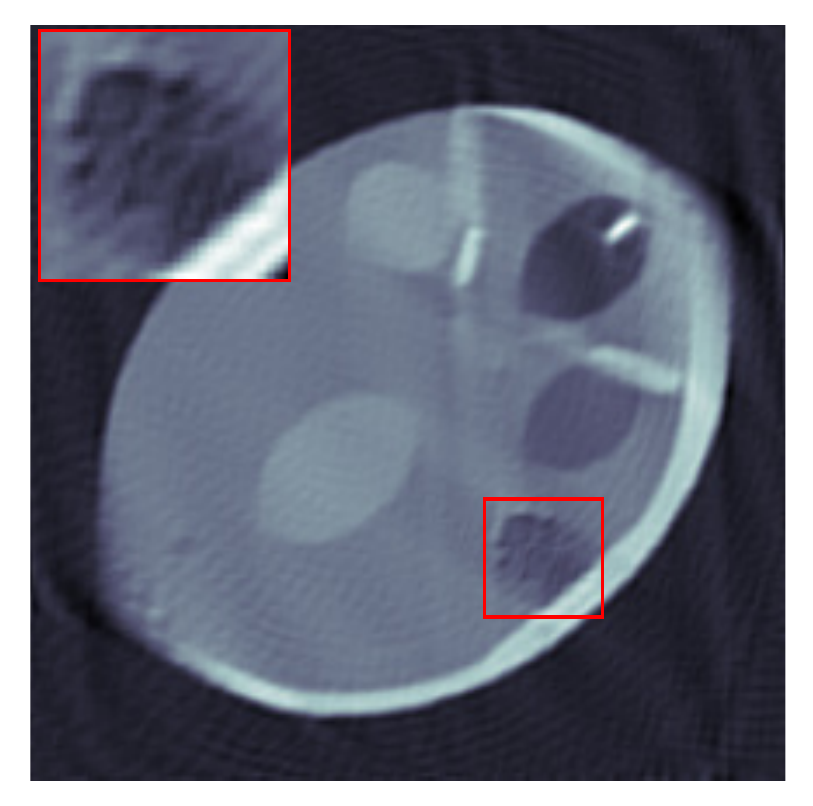}
\label{fig:CT pseudo inverse}}
\subfloat[$\Po_\solspace(\signal_{\Ro,1})$]{\includegraphics[width=0.3\textwidth]{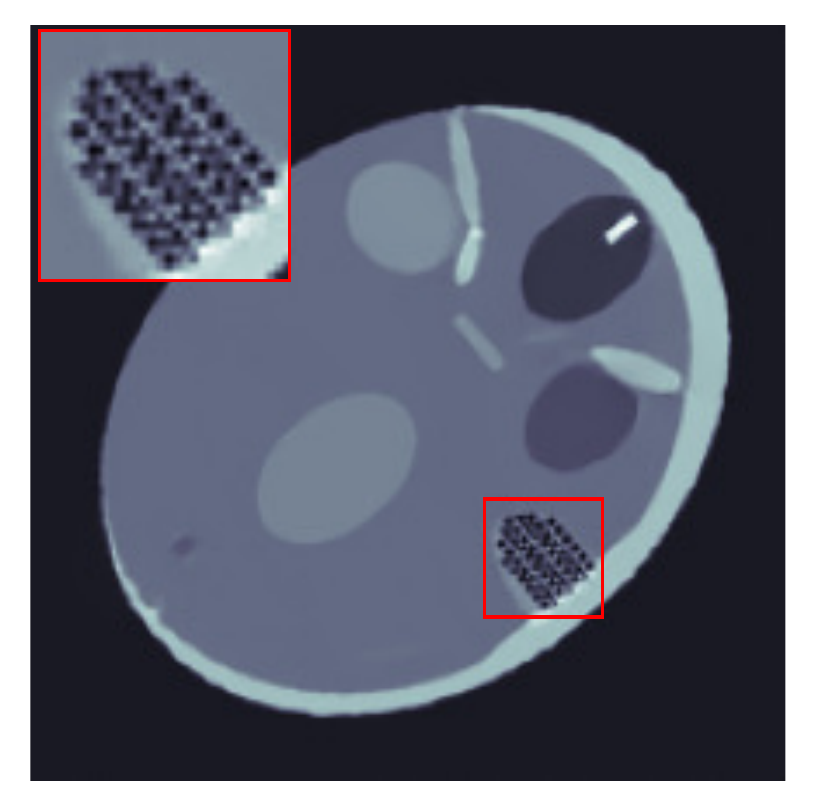}
\label{fig:CT cascaded resnet}}\\ 
\vspace{-.5em}
\subfloat[$\Po_{\solspace} (\signal_{\Ro,2})$]{\includegraphics[width=0.3\textwidth]{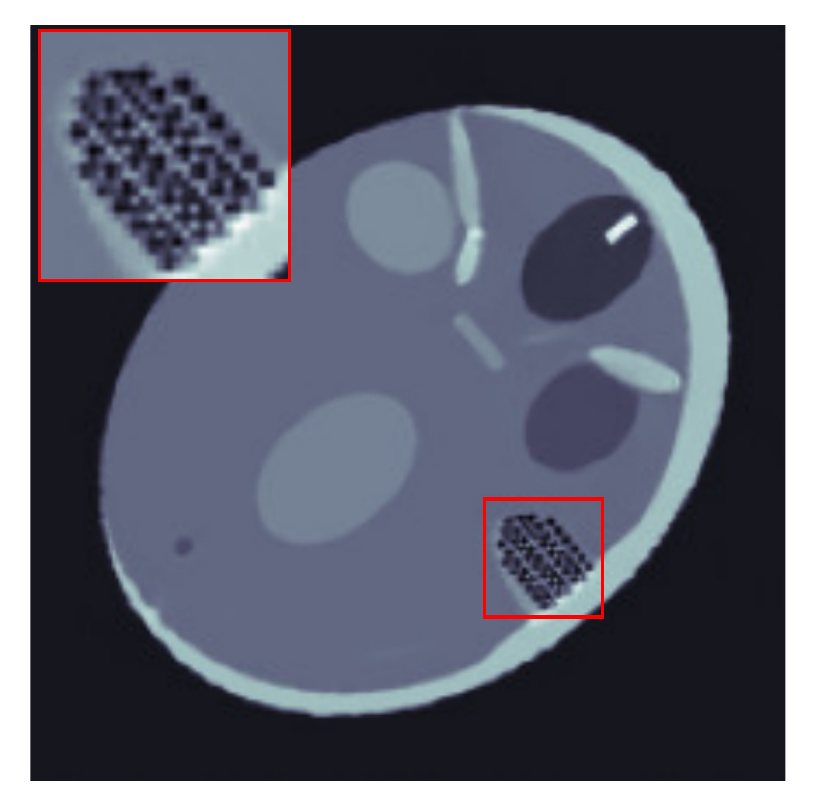}
\label{fig:CT iter cascaded unet}}
\subfloat[$\signal_{\Psi,1}$]{\includegraphics[width=0.3\textwidth]{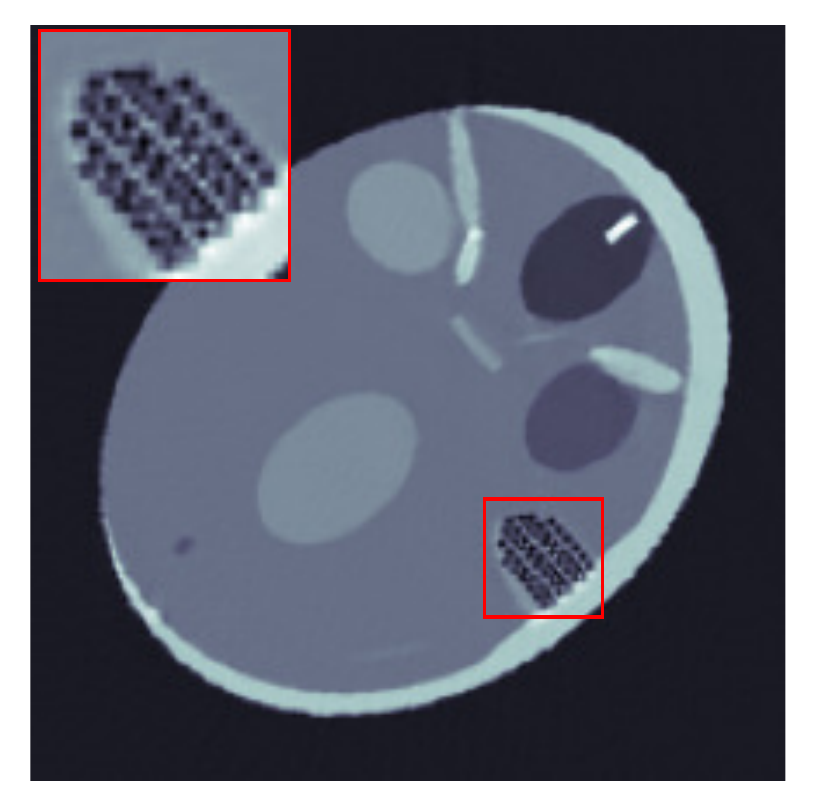}
\label{fig:CT null space}}
\subfloat[$\signal_{\Psi,2}$]{\includegraphics[width=0.3\textwidth]{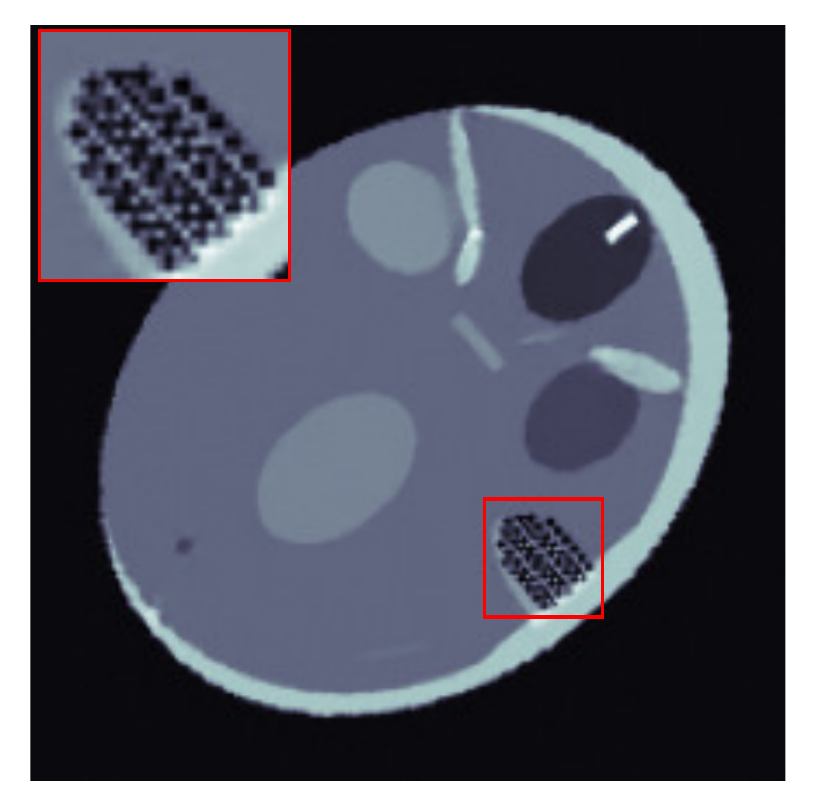}
\label{fig:CT cascaded null space}}
\caption{\textbf{Limited view CT reconstruction results}. Ground truth image $x_0$  (randomly selected),  pseudoinverse reconstruction $\signal^\ddag$, projected two-step reconstructions $\Po_{\solspace} \paren{\signal_{\Ro,1}}$, $\Po_{\solspace} \paren{\signal_{\Ro,2}}$ and null-space reconstructions $\signal_{\Psi,1}$, $\signal_{\Psi,2}$.
}
\label{fig:ct-results}
\end{figure}

\subsection{Study B: limited angle CT}

In our second study, we address limited angle CT, where the angular range of the available data is limited to a strict subset. The experiments are based on artificial phantoms. The structure of the phantoms $(\signal_i)_{i=1}^N$ of size $192 \times 192$ pixels is based on the Shepp-Logan phantom. Each sample consists of a randomly rotated phantom disc that contains smaller ellipses and rectangles drawn in a random fashion. Furthermore, some instances also contain high-frequency details (cf. Figure \ref{fig:ct-results}). All in all we generated 2000 toy samples for training and another 200 slices for testing.

\subsubsection*{Implementation details}
	
	The forward operator takes the form $\Ao = \So\circ \Ro$, where $\Ro$ is the discretized Radon transform with full angular range $[0^\circ,180^\circ)$. Here, $\So\in\{0,1\}^{m\times m }$ is a binary mask for removing angular projections such that we obtain a total of $60$ angular projections with angular directions in $[0^\circ,120^\circ]$. We again make use of the Moore-Penrose inverse as initial reconstruction method. In this case, the inverse is approximated by the singular value decomposition (SVD) \cite{engl1996}. Python code for  the discretization of $\Ao$ and a stable inexact   pseudoinverse using truncated SVD is provided in our github repository \url{https://github.com/anger-man/cascaded-null-space-learning}.

	\begin{table}[htb!]
		\normalsize
		\begin{center}
			\caption{{\textbf{Limited view CT}. Quantitative evaluation on 200 test phantoms. The reported metrics are $\psnr$ and SSIM$\times 100$ (higher is better).}}
			\begin{tabular}{l | c | c | c|c }
				\toprule
				\multicolumn{5}{c}{\textbf{benchmark}}\\ \midrule
				& $\signal_{\Ro,1}$ & $\Po_\solspace(\signal_{\Ro,1})$ & $\signal_{\Ro,2}$  &$\Po_\solspace(\signal_{\Ro,2})$\\ \midrule
				$\psnr$ &34.88&34.9&35.97&35.98 \\ \midrule
				$100 \times \ssim$ &98.38&98.36&98.71&98.7 \\ \midrule
				\multicolumn{5}{c}{\textbf{null space networks}}\\ \midrule
				&$\signal_{\Psi,1}$& $\signal_{\Psi,1}^\text{unc}$ 	&$\signal_{\Psi,2}$& $\signal_{\Psi,2}^\text{unc}$\\ \midrule
				$\psnr$ &35.87&35.8&\textbf{37.87}&{37.14 }\\ \midrule
				$100 \times \ssim$ &97.47&97.59&\textbf{98.78}&{98.68} \\ \bottomrule
			\end{tabular}
			\label{tab:radon}
		\end{center}
		
	\end{table}

\subsubsection*{Reconstruction results}

Quantitative results are summarized in Table~\ref{tab:radon}. The baseline method $\signal_{\Ro,1}$ yields a $\psnr$ of 34.88 and SSIM of 0.984. Both metrics increase to 35.97 and 0.987 if using the cascaded U-net architecture $\signal_{\Ro,2}$. Interestingly, subsequent projection onto $\solspace$ for $\signal_{\Ro,1}$ and $\signal_{\Ro,2}$ does not enhance results at all. This is mainly due to the small stepsize of $\lambda = 0.003$ for the Landweber iterations. However, increasing the stepsize leads to instabilities and thus insufficient solutions. Again, leveraging the methods $\signal_{\Ro,1}$ and $\signal_{\Ro,2}$ to the null space schemes $\signal_{\Psi,1}$ and $\signal_{\Psi,2}$ significantly increases the $\psnr$. Best results have been achieved by the cascaded null space network $\signal_{\Psi,2}$, which yields a $\psnr$ and SSIM of 37.87 and 0.988, respectively. Better reconstruction performance is only observed in the $\psnr$ while the SSIM does not seem to improve. However,  for the toy dataset, the SSIM can be observed to already be rather small. Analogously to study A,  incorporating the uncertainty-aware risk functional in \eqref{eq:riskns} does not show any impact on the recovery performance of the null space networks.

In Figure~\ref{fig:ct-results}(b) we see the characteristic limited angle streak artifacts by in the pseudoinverse reconstruction. Again, all learned reconstruction strategies are able to remove these artifacts completely. The single U-net architectures shown in Figure \ref{fig:ct-results} (c) and (e) have problems when reconstructing high-frequency details. This can be clearly observed when we look at the enlarged patches in the red boxes. Qualitative reconstruction of these patches is more satisfying for the cascaded architectures in Figure~\ref{fig:ct-results}~(d) and (f), while the latter yields the best visual  reconstruction compared to the ground truth.

\subsubsection*{Uncertainty quantification}
\begin{figure}[htb!]
\centering
\includegraphics[width=0.32\textwidth]{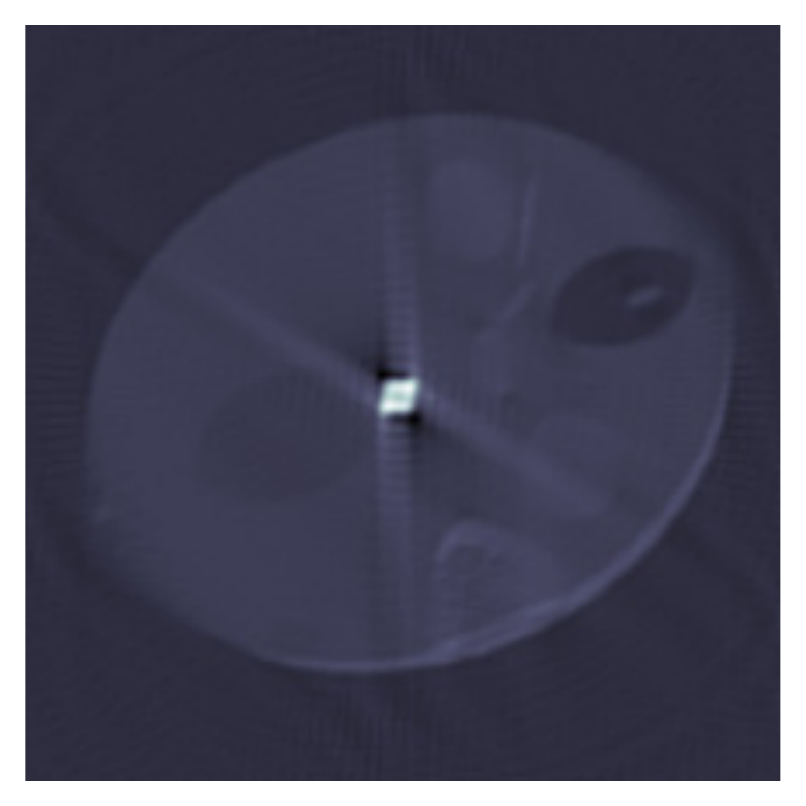}
\includegraphics[width=0.32\textwidth]{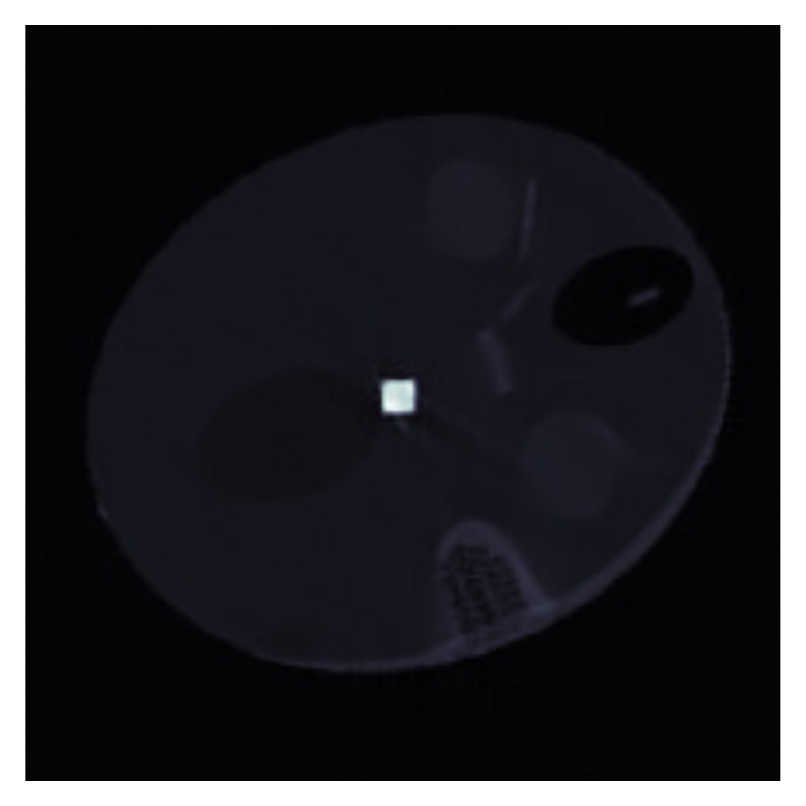}
\includegraphics[width=0.32\textwidth]{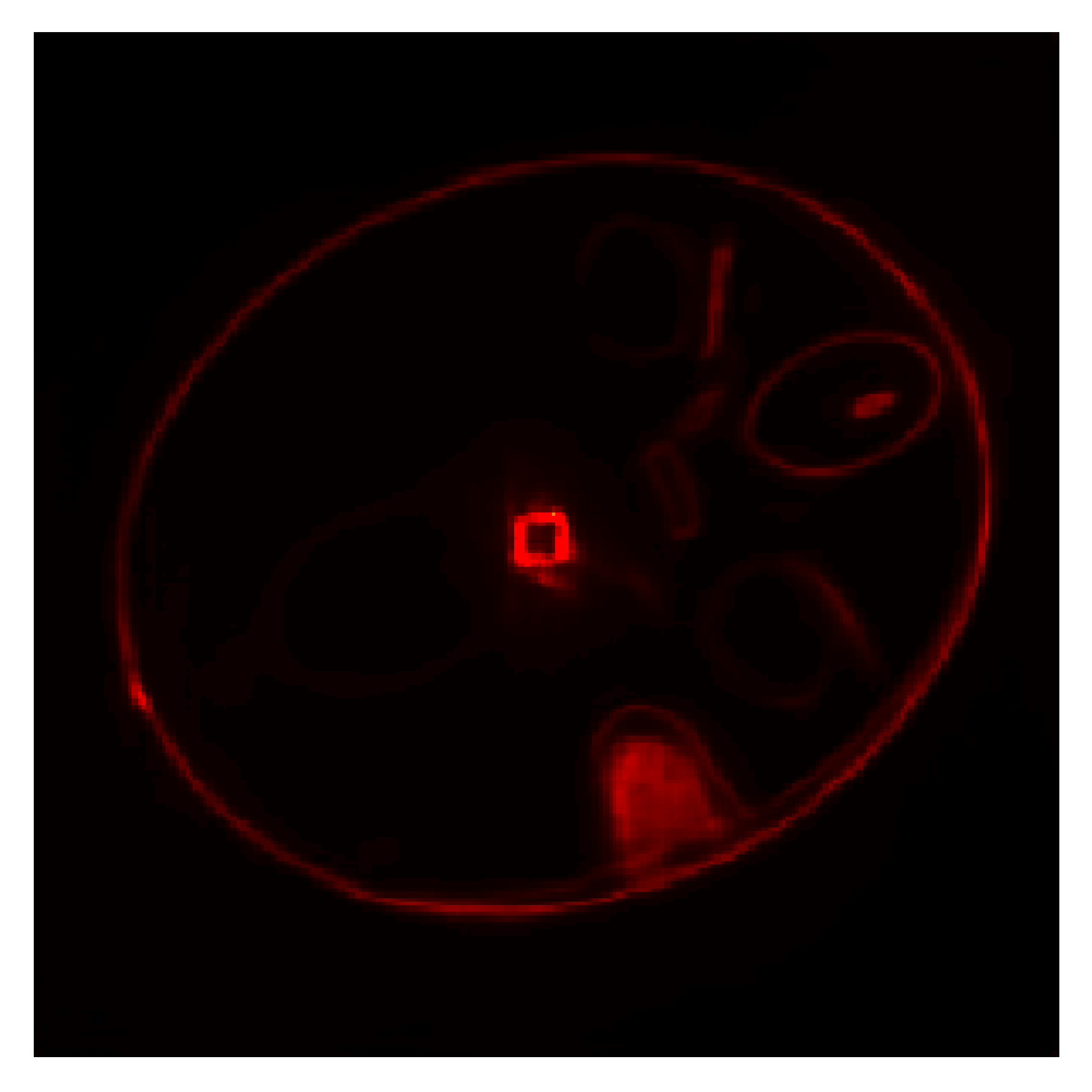}
\\
\includegraphics[width=0.32\textwidth]{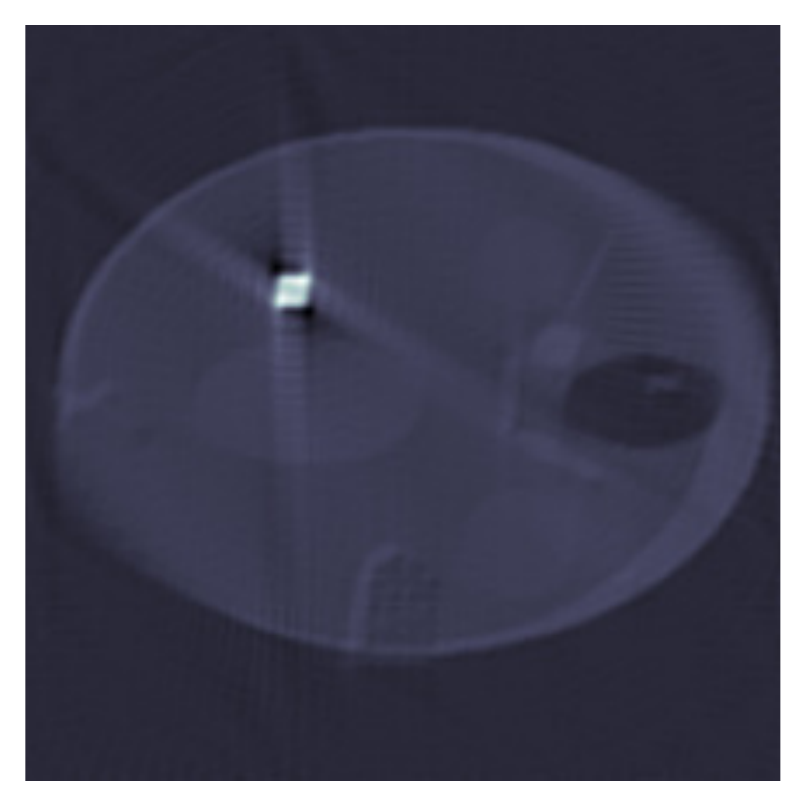}
\includegraphics[width=0.32\textwidth]{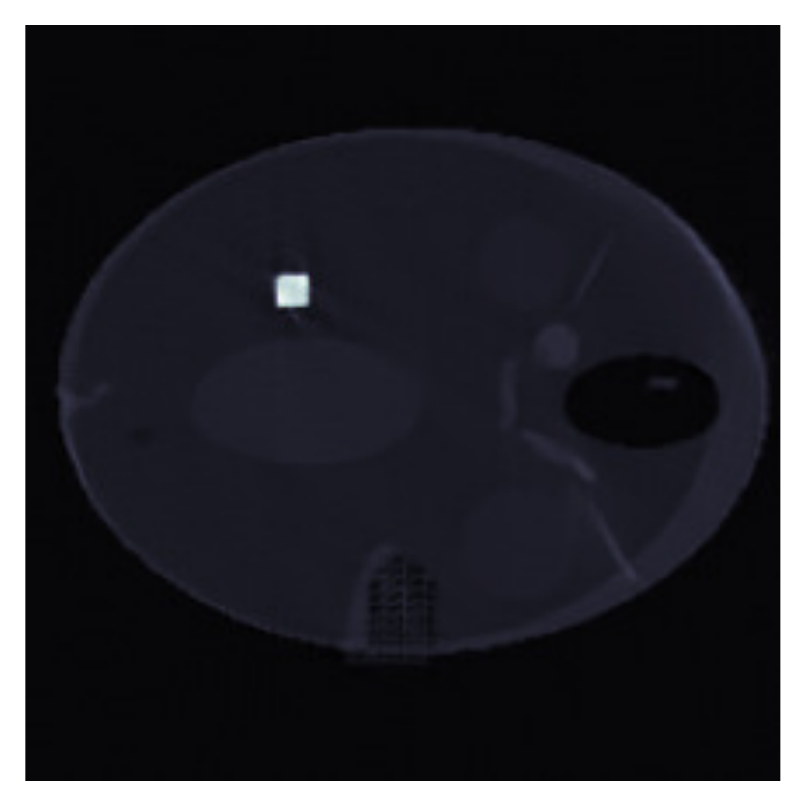}
\includegraphics[width=0.32\textwidth]{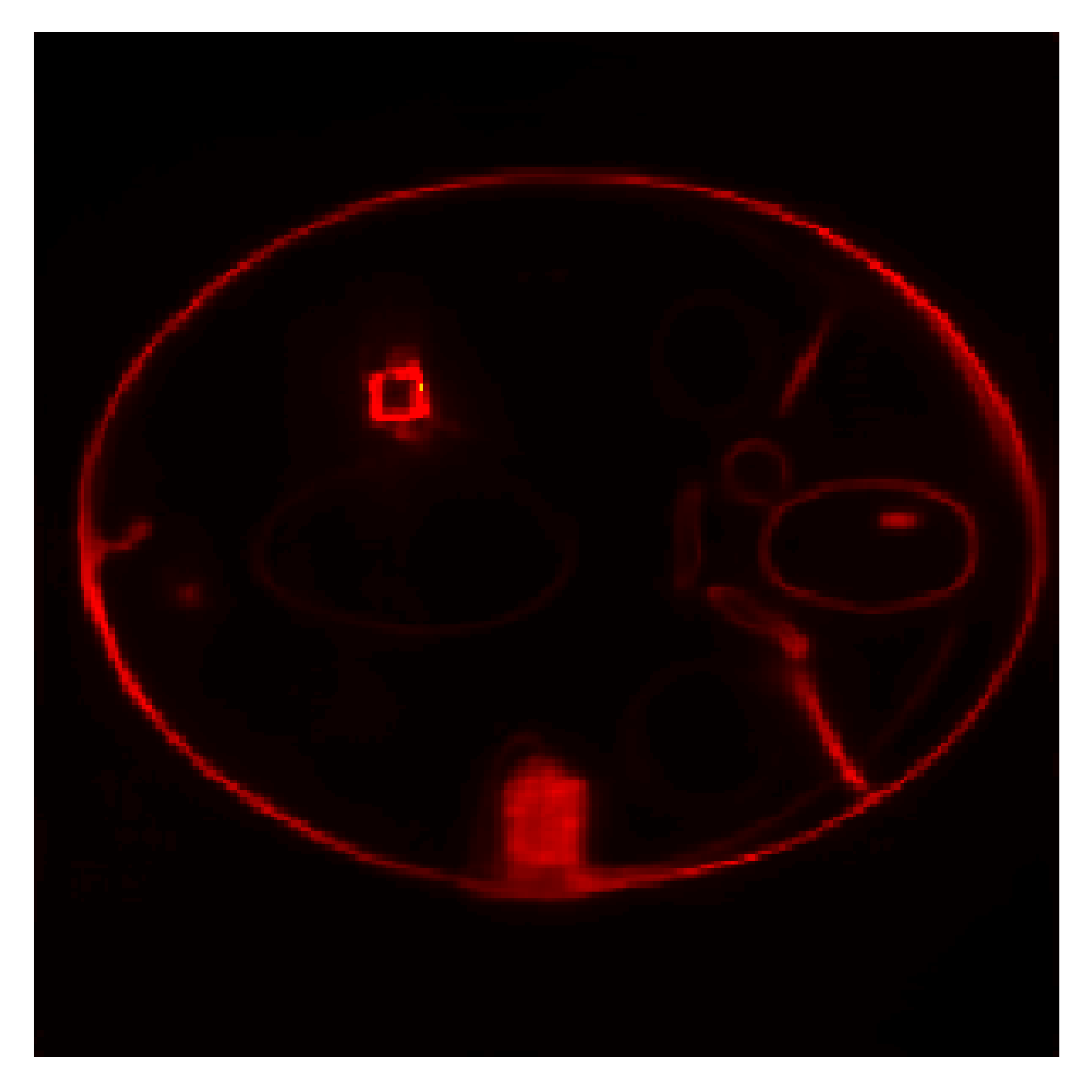}
\caption{\textbf{Position-related uncertainty in limited view CT}. 
The presence of an unknown material in the synthetic phantomcauses artifacts classical reconstruction $x^\ddag$ (first column). While the reconstruction method $\Psi_\theta^{(2)}(x^\ddag)$ reliably reconstructs the phantom (second column) the predicted uncertainty maps (third column) clearly identifies this irregular object and  unreliable regions in the prediction by a higher uncertainty values.}
\label{fig:radon_unc}
\end{figure}	

Similar to Study A, the benefits of simultaneously modeling the uncertainty maps for the toy phantom dataset are qualitatively explored. In this case, we simulate OOD data by adding a square object to the phantom. This may be interpreted as a very simple simulation of for example metal being present inside the human body, which can be the case due to dentures or artificial joints. Metal artifacts in CT can be particularly problematic in the areas surrounding the metal objects, where the image quality can be severely degraded. Again, behavior of the  Moore-Penrose inverse and the cascaded null space network is investigated on two test phantoms shown in Figure \ref{fig:radon_unc}. The inserted square perturbation yields strong streak artifacts for the model-based reconstruction $x^\ddag$ (first row) and is visible in the DL-based reconstruction of $\Psi_\theta^{(2)}(x^\ddag)$ (second row). The position of the square is clearly marked with high intensity in the corresponding uncertainty maps (third row). Also in the setting of limited angle CT the uncertainty-aware null space network clearly identifies unreliable regions during reconstruction caused by presence of unknown metal artifacts.

\section{Conclusion}
	
In this paper we presented a learning-based method for simultaneous image  reconstruction and uncertainty estimation. The proposed uncertainty estimation procedure introduces a second output branch into the reconstruction network, which can be seen as scale map for the Laplace distributed residuals between reconstructed image and ground truth. This leads to small uncertainty values for regions that are easy to learn based on the training data. On the other hand, rapid changes in pixel intensities and out of distribution data create new regions of high frequency detail, a property for which the second output branch has already been developed to predict high uncertainty. Our experiments demonstrate that simple modifications to the network architecture and slight modifications to the risk functional yield uncertainty information and increased reconstruction quality. The approach is implemented and evaluated based  on the standard U-net architecture commonly used in image processing tasks. However, our framework  can also be combined with more complex architectures or integrated into state-of-the-art image reconstruction methods.

\section*{Acknowledgment}

The contribution of C.~A. is supported by VASCage -- Centre on Clinical Stroke Research. VASCage is a COMET Centre within the Competence Centers for Excellent Technologies (COMET) programme and funded by the Federal Ministry for Climate Action, Environment, Energy, Mobility, Innovation and Technology, the Federal Ministry of Labour and Economy, and the federal states of Tyrol, Salzburg and Vienna. COMET is managed by the Austrian Research Promotion Agency (Österreichische Forschungsförderungsgesellschaft). The contribution of S.~G. is part of a project that has received funding from the European Union’s Horizon 2020 research and innovation program under the Marie Sk\l{}odowska-Curie grant agreement No. 847476. The views and opinions expressed herein do not necessarily reflect those of the European Commission.

%

\end{document}